\documentclass{article}
\usepackage{PRIMEarxiv}

\usepackage[utf8]{inputenc} 
\usepackage[T1]{fontenc}    
\usepackage{hyperref}       
\usepackage{url}            
\usepackage{booktabs}       
\usepackage{amsfonts}       
\usepackage{nicefrac}       
\usepackage{microtype}      
\usepackage{lipsum}
\usepackage{fancyhdr}       
\usepackage{graphicx}       
\graphicspath{{media/}}     
 \usepackage{natbib}
\usepackage{mathtools}
\usepackage{pdflscape}

\usepackage{amsmath}
\usepackage{amsthm}  
\usepackage{multirow}

\usepackage{pifont}
\usepackage{newunicodechar}
\usepackage{tabularx}
\usepackage{float}
\usepackage{longtable}
\usepackage{subcaption}

\newunicodechar{✓}{\ding{51}}
\newunicodechar{✗}{\ding{55}}

\newtheorem{proposition}{Proposition}
\newtheorem{definition}{Definition}
\newtheorem{theorem}{Theorem}

\usepackage{chngcntr}
\counterwithin{theorem}{section}
\counterwithin{proposition}{section}
\counterwithin{lemma}{section}
\counterwithin{corollary}{section}

\newcommand{\TV}{\operatorname{TV}}           
\newcommand{\JS}{\operatorname{JS}}           
\newcommand{\KL}{\operatorname{KL}}           
\newcommand{\Hell}{\mathrm{H}}                
\newcommand{\DeltaBF}{\Delta_{\!\mathrm{BF}}} 
\newcommand{\E}{\mathbb{E}}

\title{Process-Tensor Tomography of SGD: Measuring Non-Markovian Memory via Back-Flow of Distinguishability
\thanks{\textit{\underline{Citation (\textbf{to appear in})}}: 
\textbf{Vasileios Sevetlidis, and George Pavlidis.
Process-Tensor Tomography of SGD: Measuring Non-Markovian Memory via Back-Flow of Distinguishability. Proceedings of The 29th International Conference on Artificial Intelligence and Statistics, in Proceedings of Machine Learning Research. 2026}}
}

\author{
  Vasileios Sevetlidis* \\
  Athena RC \\
  University Campus Kimmeria \\
  Xanthi, GR67100, Greece\\
  \texttt{vasiseve@athenarc.gr} \\
   \AND
  George Pavlidis \\
  Athena RC \\
  University Campus Kimmeria \\
  Xanthi, GR67100, Greece\\
  \texttt{gpavlid@athenarc.gr} \\
}

\begin{document}
\maketitle

\begin{abstract}
This work proposes neural training as a \emph{process tensor}: a multi-time map that takes a sequence of controllable instruments (batch choices, augmentations, optimizer micro-steps) and returns an observable of the trained model. Building on this operational lens, we introduce a simple, model-agnostic witness of training memory based on \emph{back-flow of distinguishability}. In a controlled two-step protocol, we compare outcome distributions after one intervention versus two; the increase $\Delta_{\!\mathrm{BF}} = D_2 - D_1>0$ (with $D\in\{\mathrm{TV}, \mathrm{JS}, \mathrm{H}\}$ measured on softmax predictions over a fixed probe set) certifies non-Markovianity. We observe consistent positive back-flow with tight bootstrap confidence intervals, amplification under higher momentum, larger batch overlap, and more micro-steps, and collapse under a \emph{causal break} (resetting optimizer state), directly attributing the effect to optimizer/data-state memory. The witness is robust across TV/JS/Hellinger, inexpensive to compute, and requires no architectural changes. We position this as a \emph{measurement} contribution: a principled diagnostic and empirical evidence that practical SGD deviates from the Markov idealization. An exploratory case study illustrates how the micro-level signal can inform curriculum orderings. “Data order matters” turns into a testable operator with confidence bounds, our framework offers a common stage to compare optimizers, curricula, and schedules through their induced training memory.\end{abstract}

\keywords{training dynamics, process tensor, non-Markovianity, stochastic gradient descent, back-flow, distinguishability, momentum, causal break, curriculum learning, probe set}

\section{INTRODUCTION}

Modern training pipelines make---and often quietly rely on---simplifying assumptions about how updates compose over time. In particular, theoretical analyses frequently model training as (approximately) Markovian once the algorithm’s internal buffers are included in the state: a future iterate is conditionally independent of earlier choices  given current parameters and optimizer state. This idealization keeps proofs tractable,  but it leaves practitioners without a direct, operational way to \emph{measure} whether a real training run is memoryless or history-dependent. Empirical analyses of reshuffling and curricula confirm that order matters, and optimizer designs intentionally carry  history, yet there is no simple diagnostic that a practitioner can run to falsify a ``memoryless training'' hypothesis~\citep{shamir2016without,gurbu2015rrbeats,sutskever2013importance,goh2017momentum,even2023markovsgd}. A concrete pain point is schedule design (data order, augmentations, and optimizer knobs): many heuristics work, yet we lack a principled diagnostic to tell when past choices still influence today’s 
outcomes~\citep{bottou2018optimization,mandt2017sgd_sde}.

We build on operational notions of non-Markovianity developed in open quantum systems. There, increases in distinguishability between states (``information back-flow'') certify memory effects~\citep{breuer2009measure,rivas2014review}. The \emph{process tensor} formalism elevates this to a full multi-time description: dynamics are maps from sequences of controllable \emph{instruments} to outcome distributions, with a sharp \emph{operational Markov condition} and the notion of a \emph{causal break}~\citep{pollock2018process_tensor,pollock2018operational_markov}. Comprehensive reviews and tomographic schemes further consolidate this perspective~\citep{milz2021review,white2022nMQPT}. In our classical setting, we treat training as a process tensor: a multi-time map from chosen instruments (batch selection, augmentation, optimizer micro-step) to observables (predictions on a probe set). We quantify distinguishability using TV/JS/Hellinger metrics on model outputs, and we implement a causal break by \emph{resetting optimizer state}. Adapting the standard witness from open-systems theory, we test for \emph{back-flow of distinguishability}: run two one-step histories that differ only in the first intervention $A$ vs.\ $A'$, then apply a common second intervention $B$. If the distance between resulting prediction distributions increases after $B$, we certify memory (non-Markovianity). This transposes the trace-distance back-flow criterion and process-tensor framework into the context of SGD.

\textbf{Contributions.}
\begin{enumerate}
  \item \textbf{Operational witness for SGD memory.} We introduce a two-step A/B protocol and a \emph{back-flow} statistic computed on predictive distributions (TV/JS/Hellinger). The method is measurement-only and model-agnostic.
  \item \textbf{Mechanism test via \emph{causal break}.} We instantiate a ``break'' (reset optimizer state) before $B$. If optimizer buffers mediate memory, $\Delta$ should collapse---a direct, falsifiable mechanism check~\citep{pollock2018operational_markov}.
  \item \textbf{Empirical \emph{process-tomography} of training.} On CIFAR-10 and Imagenette across multiple architectures and three divergences, we measure $\Delta$ under controlled regimes (standard, resonant, orthogonal, negative) and report seed-robust CIs.
  \item \textbf{Ablations that map the control parameters.} We chart how $\Delta$ scales with momentum, batch overlap, and step depth; and we quantify the collapse under causal break.
\end{enumerate}

The structure of this work is as follows: background and related work is presented in \S\ref{sec:rw}; \S\ref{sec:method} formalizes the process-tensor view and the back-flow witness; \S\ref{sec:procedure} details the A/B protocol and the causal-break mechanism test; \S\ref{sec:results} reports measurements and ablations. Code, logs, and plotting scripts accompany the paper for full reproducibility.

\section{RELATED WORK}\label{sec:rw}

Analyses of stochastic optimization often treat updates as (time-homogeneous) Markov dynamics once one augments the state with algorithmic buffers and assumes i.i.d.\ sampling. This viewpoint underpins tutorials and SDE-based models of SGD: it enables mixing/ergodicity arguments and stationary-distribution pictures near minima~\citep{bottou2018optimization,mandt2017sgd_sde}. Sampling \emph{without} replacement (random reshuffling) couples consecutive steps through the epoch permutation and changes both analysis and practice; multiple works prove or quantify RR’s advantages and make its temporal dependence explicit. These effects can be re-cast as Markov only by further augmenting the state (e.g., with a permutation pointer), underscoring that order is a first-class control knob in modern training~\citep{shamir2016without,rajput2020rr,gurbu2015rrbeats}. Momentum and adaptive methods explicitly accumulate history. Classic and modern accounts (heavy-ball, \citep{nesterov1983}; \citep{sutskever2013importance}; \citet{goh2017momentum}’s Distill note) document their empirical and dynamical impact; stochastic modified-equation analyses encode them as SDEs in an \emph{enlarged} state. Our \emph{causal break} experiment operationalizes this intuition: if buffers mediate history, resetting them should null out back-flow---which is exactly what we observe~\citep{polyak1964,li2017sme}.

A separate line treats the \emph{data source} as Markov and tracks the role of mixing time in SGD’s convergence---formally moving dependence from the algorithm to the sampler. This is adjacent to, but distinct from, our aim: we test for history-dependence in the \emph{training process} as exposed by output distributions under controlled interventions~\citep{even2023markovsgd,doan2020markovsgd,dorfman2022dependent}. Empirical work finds that gradient noise during deep learning is heavy-tailed and anisotropic, challenging Gaussian, white-noise assumptions common in SDE models. While not phrased in ``Markov vs.\ non-Markov'' terms, these observations motivate measurement-first diagnostics that do not hinge on specific noise models~\citep{simsekli2019heavytailed,zhu2019anisotropic}. Several diagnostics quantify how predictions or representations change across training. LOOD-type measures track output changes induced by adding/removing examples (aimed at privacy/memorization)~\citep{ye2023lood,carlini2019secret,carlini2021extract,koh2017influence,pruthi2020tracin,yeh2018representer}, while representation-similarity metrics (e.g., CKA) probe internal drift~\citep{kornblith2019cka}. Recent work questions their reliability and sensitivity~\citep{davari2022cka_reliability}. In contrast, our witness is \emph{non-parametric, intervention-defined}, and explicitly tied to an operational Markov condition.

\section{Methodology}\label{sec:method}

We formalize a short segment of training as a multi-time stochastic process that maps \emph{instruments}---controllable, randomized interventions---to a distribution over \emph{observables} on a fixed probe set. The key object is a classical \emph{process tensor} (comb) acting at two times; our operational witness of memory is a \emph{back-flow of distinguishability} between two one-step histories under a fixed second intervention. We prove that a positive back-flow contradicts an \emph{operational Markov} condition at the observable, and we show that a \emph{causal break} (resetting optimizer buffers) restores that condition under a mild sufficiency hypothesis, predicting the collapse of our witness.

\subsection{State spaces, instruments, and kernels}\label{sec:spaces}
Let $(\mathcal{S},\Sigma_{\mathcal{S}})$ be a standard Borel space for the \emph{latent algorithmic state} (parameters and any optimizer buffers), and $(\mathcal{O},\Sigma_{\mathcal{O}})$ an observable space (per-example class probabilities on a fixed probe set). A \emph{stochastic kernel} $K:\mathcal{A}\times\Sigma_{\mathcal{B}}\to[0,1]$ maps each $a\in\mathcal{A}$ to a probability measure on $(\mathcal{B},\Sigma_{\mathcal{B}})$ and is measurable in $a$.

An \emph{instrument} $I$ specifies: (i) a mini-batch index set, (ii) a data-augmentation kernel, and (iii) optimizer micro-parameters used for $k$ micro-steps. Executing $I$ induces a latent kernel $\mathcal{K}_I:\mathcal{S}\leadsto\mathcal{S}$ (the random update over $k$ micro-steps). Given two successive instruments $I_0,I_1$ and an initial prior $\pi_0\in\mathcal{P}(\mathcal{S})$, the latent two-time law is
\begin{equation}\label{eq:latent-two-time}
\pi_2^{(I_0,I_1)}(\mathrm{d}s_2)
\;=\; \int_{\mathcal{S}} \!\!\Bigg[\int_{\mathcal{S}}
\mathcal{K}_{I_1}(s_1,\mathrm{d}s_2)\;\mathcal{K}_{I_0}(s_0,\mathrm{d}s_1)\Bigg]\;\pi_0(\mathrm{d}s_0).
\end{equation}
The \emph{observation channel} $\mathcal{O}:\mathcal{S}\leadsto\mathcal{O}$ pushes latent state to predictions on the probe. The observable one- and two-time laws are then
\begin{align}
\Phi_1^{(I_0)}(\mathrm{d}o) &= \int_{\mathcal{S}} \mathcal{O}(s_1,\mathrm{d}o)\,\pi_1^{(I_0)}(\mathrm{d}s_1),\\
\Phi_2^{(I_0,I_1)}(\mathrm{d}o) &= \int_{\mathcal{S}} \mathcal{O}(s_2,\mathrm{d}o)\,\pi_2^{(I_0,I_1)}(\mathrm{d}s_2),
\end{align}
where $\pi_1^{(I_0)} = \pi_0\mathcal{K}_{I_0}$ and composition is the usual \emph{link product} of kernels: $(K_2\star K_1)(a,\mathrm{d}c):=\int K_2(b,\mathrm{d}c)\,K_1(a,\mathrm{d}b)$.

Collecting all two-time statistics defines a multilinear map
\[
\mathsf{T}_{2{:}0}: \;\mathfrak{I}_0\times\mathfrak{I}_1\to \mathcal{P}(\mathcal{O}),\qquad
(I_0,I_1)\mapsto \Phi_2^{(I_0,I_1)},
\]
where $\mathfrak{I}_t$ is a convex set of admissible (randomized) instrument kernels. In the classical setting, complete positivity reduces to \emph{stochasticity} and \emph{non-negativity} under linking with any admissible post-processing on observables.

\subsection{Operational Markov condition at the observable}\label{sec:omc}

\begin{definition}\textbf{Operational Markov condition (OMC) at the observable}.
Fix a second-step instrument $B$. We say that the two-time process is 
\emph{operationally Markov at the observable for $B$} if there exists 
a stochastic map $\Lambda_B : \mathcal{P}(\mathcal{O}) \to \mathcal{P}(\mathcal{O})$ 
such that for every first-step instrument $I$,
\begin{equation}\label{eq:omc}
    \Phi_2^{(I,B)} \;=\; \Lambda_B\!\big[\Phi_1^{(I)}\big].
\end{equation}
\end{definition}

\begin{proposition}[Proposition 3.1]
If the two-time process is operationally Markov at the observable for $B$, 
then for any contractive divergence $D$, the back-flow of distinguishability 
satisfies $\Delta^{(I,B)} \leq 0$ for all $I$.
\end{proposition}

Intuitively, once the observed mid-time law $\Phi_1$ is fixed, $B$ acts by the \emph{same channel} on it, independent of how $\Phi_1$ was prepared. Condition~\eqref{eq:omc} is strictly weaker than assuming latent Markov dynamics in $(\theta,\text{buffers})$: it only constrains the \emph{measured} map.

\subsection{A back-flow witness of operational memory}\label{sec:witness}
Choose two first-step instruments $A,A'$ (same indices, different augmentation) and fix $B$. Let $D$ be a divergence on $\mathcal{P}(\mathcal{O})$ that obeys data processing (see \S\ref{sec:divs}). Define
\begin{equation}
D_1 := D\!\big(\Phi_1^{(A)},\Phi_1^{(A')}\big),
\end{equation}

\begin{equation}
D_2 := D\!\big(\Phi_2^{(A,B)},\Phi_2^{(A',B)}\big),
\end{equation}

\begin{equation}
\DeltaBF := D_2 - D_1.
\end{equation}

\begin{proposition}[No back-flow under OMC]\label{prop:nobackflow}
If \eqref{eq:omc} holds and $D$ satisfies data processing, then $\DeltaBF\le 0$.
\end{proposition}

Thus, \emph{positive} back-flow ($\DeltaBF>0$) \emph{falsifies} \eqref{eq:omc} and witnesses operational memory at the two-step scale. Mechanistically this can arise because (i) history is \emph{carried} across the step by unobserved buffers that modulate the action of $B$, or (ii) the observable is \emph{insufficient} for how $B$ acts: two distinct latent preparations that agree on $\Phi_1$ evolve differently under $B$.

\subsection{Causal break and a sufficient condition for restoring OMC}\label{sec:break}
A \emph{causal break} before $B$ is a kernel $\mathcal{B}:\mathcal{S}\leadsto\mathcal{S}$ that erases a memory channel we intend to cut. In our setting, it resets optimizer buffers while keeping parameters fixed: if $s=(\theta,\upsilon)$, then $\mathcal{B}(s,\mathrm{d}\theta'\mathrm{d}\upsilon')=\delta_{\theta}(\mathrm{d}\theta')\,\delta_{0}(\mathrm{d}\upsilon')$. Let $\widetilde{\mathcal{K}}_B := \mathcal{K}_B\star \mathcal{B}$ be the post-break latent kernel.

We give a mild, checkable condition under which the break implies \eqref{eq:omc}.

\begin{definition}[Observable sufficiency for $B$]\label{def:sufficient}
Let $\mathcal{Q}\subseteq\mathcal{P}(\mathcal{S})$ be the set of mid-time latent laws reachable after the first step. The observation channel $\mathcal{O}$ is \emph{sufficient for $B$ on $\mathcal{Q}$} if for any $\pi_1,\pi_1'\in\mathcal{Q}$ with equal observables $\pi_1\mathcal{O}^{-1}=\pi_1'\mathcal{O}^{-1}$, we also have $\pi_1\widetilde{\mathcal{K}}_B\mathcal{O}^{-1}=\pi_1'\widetilde{\mathcal{K}}_B\mathcal{O}^{-1}$.
\end{definition}

\begin{theorem}\textbf{(Break $\Rightarrow$ observable channel under sufficiency.)}\label{thm:break-channel}
If a causal break $\mathcal{B}$ is applied and $\mathcal{O}$ is sufficient for $B$ on $\mathcal{Q}$, then there exists a stochastic map $\Lambda_B$ on $\mathcal{P}(\mathcal{O})$ such that for all first-step instruments $I$, $\Phi_2^{(I,B)}=\Lambda_B[\Phi_1^{(I)}]$. Consequently, $\DeltaBF\le 0$ for any contractive divergence $D$.
\end{theorem}

A sufficient way to ensure Def.~\ref{def:sufficient} is a \emph{Markov factorization} through the observable: there exists a lifting kernel $R:\mathcal{O}\leadsto\mathcal{S}$ (a right-inverse of $\mathcal{O}$ on $\mathcal{Q}$) such that $\widetilde{\mathcal{K}}_B = \widetilde{\mathcal{K}}_B\star R\star \mathcal{O}$ on $\mathcal{Q}$. Then $\Lambda_B=\mathcal{O}\star \widetilde{\mathcal{K}}_B\star R$ is explicit.

\subsection{Perturbative scaling predictions}\label{sec:perturb}
A first-order linearization explains the observed trends with momentum, batch overlap, and micro-step depth. Write the $k$-step update under instrument $I$ as $U_I(\theta)\approx \theta - \eta\sum_{t=0}^{k-1}\hat{g}_I(\theta,t)$, where $\hat{g}$ incorporates momentum $\mu$ via $\hat{g}_I(\cdot,t)\approx(1-\mu)\sum_{i=0}^{t}\mu^i g_I(\cdot)$. For the two histories $A,A'$,
\[
\theta^{(A)}_1-\theta^{(A')}_1 \;\approx\; -\eta\sum_{t=0}^{k-1}\big(\hat{g}_A-\hat{g}_{A'}\big)(\theta_0,t),
\]
and after applying $B$ once more,
\begin{align*}
\theta^{(A,B)}_2 - \theta^{(A',B)}_2 
&\;\approx\; \Big(I - \eta k\,H_B(\theta_0)\Big)\big(\theta^{(A)}_1-\theta^{(A')}_1\big) \\
&\quad+\; \eta k\,\Xi_{AB}.
\end{align*}

where $H_B$ is a (preconditioned) Jacobian/Hessian of the $B$-gradient and $\Xi_{AB}$ collects curvature--noise cross-terms. Two corollaries match our measurements:
(i) \textbf{Momentum amplification:} $\|\theta^{(A)}_1-\theta^{(A')}_1\|$ scales like $(1-\mu^k)/(1-\mu)$, so $D_2{-}D_1$ increases with $\mu$.
(ii) \textbf{Resonance via overlap:} when $B$ reuses samples from $A$ (overlap $\rho$), $H_B$ and $\Xi_{AB}$ correlate with $g_A-g_{A'}$, enlarging $D_2{-}D_1$; disjoint $B$ damps it.
A causal break zeros momentum at the second step, removing the $\mu$-dependent amplification and driving $\DeltaBF\to 0$ up to $\mathcal{O}(\eta^2)$.

\subsection{Divergences and contractivity}\label{sec:divs}
We operate on per-example predictive distributions over the probe set and aggregate by averaging the per-example divergence.\footnote{Formally, if $x\in\mathcal{P}$ ranges over probe inputs and $P_x,Q_x$ are the class-probability vectors under two histories, we use $\bar{D}(P,Q):=\frac{1}{|\mathcal{P}|}\sum_{x\in\mathcal{P}} D(P_x,Q_x)$. If $D$ obeys data processing pointwise, so does $\bar{D}$.}
We require boundedness in $[0,1]$, sensitivity to multi-class changes, and data processing:
\emph{total variation} $\mathrm{TV}(p,q)=\tfrac12\|p-q\|_1,$ and \emph{Hellinger} $\mathrm{Hell}(p,q)=\tfrac12\left\|\sqrt{p}-\sqrt{q}\right\|_2$ are $f$-divergences and thus contractive under stochastic maps; \emph{Jensen--Shannon} $\mathrm{JS}(p,q)=\tfrac12\mathrm{KL}(p\|m)+\tfrac12\mathrm{KL}(q\|m),~m=\tfrac12(p+q)$ is a symmetrized mixture of $\KL$ and inherits the same property. Proposition~\ref{prop:nobackflow} therefore applies to all three.

\subsection{Inference at the observable level}\label{sec:inference}
Each run produces random variables $D_1, D_2\in[0,1]$ and $\DeltaBF\in[-1,1]$. We treat the \emph{operational Markov} inequality as the null $H_0:\E[\DeltaBF]\le 0$. We report nonparametric bootstrap CIs for the mean and perform TOST (\emph{two one-sided tests}) against a small equivalence margin $\varepsilon$ to declare practical nullity. A lower confidence bound strictly above $0$ rejects $H_0$ and certifies operational non-Markovianity for the specified instruments and observable; collapse toward $0$ after a causal break supports the mechanism diagnosis in \S\ref{sec:break}.

\subsection{What a positive back-flow \emph{means} (and what it does not)}\label{sec:meaning}
A strictly positive $\DeltaBF$ says there is \emph{no} single channel $\Lambda_B$ that maps \emph{the measured} one-step laws to the two-step laws for the fixed $B$. It \emph{does not} claim the augmented latent dynamics $(\theta,\text{buffers})$ are non-Markov; rather, it pinpoints an observable-level failure of \eqref{eq:omc}. Our causal-break experiments remove optimizer memory as a carrier and are consistent with the picture that, at the two-step scale we probe, buffers mediate the observed dependence.

\section{EXPERIMENTAL PROCEDURE}
\label{sec:procedure}

\begin{table*}[h]
\centering
\caption{Micro-step regimes.}
\label{tab:regimes}
\renewcommand{\arraystretch}{1.05}
\begin{tabular}{l c c c c c c c}
\toprule
Regime & $k$ & LR & mom & $\text{aug}_A$ & $\text{aug}_{A'}$ & $\text{aug}_B$ & overlap / classes \\
\midrule
\texttt{standard}         & 3 & 0.02 & 0.90 & weak  & color & weak & $0.5$ / \texttt{True}\\
\texttt{resonant strong} & 6 & 0.03 & 0.99 & color & blur  & weak & $1.0$ / \texttt{True}\\
\texttt{resonant mid}    & 6 & 0.03 & 0.95 & color & blur  & weak & $0.75$ / \texttt{True}\\
\texttt{orthogonal}       & 6 & 0.03 & 0.99 & color & blur  & blur & $0.0$ / \texttt{False}\\
\texttt{negative} (ctrl)  & 1 & 0.005& 0.00 & none  & none  & none & $0.0$ / \texttt{False}\\
\bottomrule
\end{tabular}
\end{table*}

We instantiate the two-step $A{\to}B$ protocol from \S\ref{sec:method} across datasets, model families, and micro-step “regimes,” and measure the observable-level back-flow $\DeltaBF = D_2 - D_1$ using $\TV$ (primary) and $\JS$/$\Hell$ (secondary). The only conceptual manipulation relative to \S\ref{sec:method} is the \emph{causal break}: either we carry the optimizer buffers from $A$ into $B$ (\emph{no break}) or we re-initialize the optimizer immediately before $B$ (\emph{break}), holding hyperparameters fixed. Everything else below specifies what each symbol means in practice.

We evaluate on \texttt{CIFAR-100} and \texttt{Imagenette}. Inputs are resized to $32{\times}32$ (RGB). For Imagenette we use the official validation split as held-out data. Backbones: \texttt{SmallCNN}, \texttt{ResNet-18} (CIFAR stem), \texttt{VGG-11}, \texttt{MobileNetV2}, and a \texttt{ViT-B/16} configured for $32{\times}32$ (2$\times$2 patches). The final linear heads match the dataset’s number of classes.

Unless otherwise stated, we use the \texttt{early} stage: SGD (LR $0.1$, momentum $0.9$, weight decay $5{\times}10^{-4}$), cosine schedule over $3$ epochs with weak CPU-side augmentation (random crop $32$ with pad $4$ and horizontal flip). We also retain \texttt{init} (random) and \texttt{late} ($20$ epochs) for sensitivity checks. A GPU snapshot of the base parameters is cached to enable fast, in-place resets between micro-experiments.

\subsection{Micro-step protocol ($A$ then $B$)}
\label{sec:micro}
A single \emph{micro-experiment} starts from the cached base parameters, applies $k$ SGD steps on a mini-batch stream from $A$, then $k$ steps from $B$. Unless otherwise specified: batch size $256$, per-step weight decay $10^{-4}$, gradient clip at $1.0$, channels-last memory layout, and \emph{AMP disabled} (the code path supports it but we keep it off in the sweep). BatchNorm running statistics are frozen during micro-steps.

We implement GPU-native transforms:
\emph{weak} = random crop ($32$, reflect pad $4$) + horizontal flip ($p{=}0.5$);
\emph{color} = weak + color jitter (brightness/contrast/saturation $0.4$) + random grayscale ($p{=}0.2$);
\emph{blur} = weak + depthwise Gaussian blur (kernel $3$, $\sigma\!\sim\!\mathcal{U}[0.1,2.0]$);
and \emph{none} = identity.

For each repeat we draw $I_A$ uniformly without replacement (size $256$) from cached training tensors.
We draw $I_B$ to achieve a prescribed overlap with $I_A$,
$|I_A\cap I_B|\approx\lfloor \texttt{overlap}\times256\rfloor$,
and optionally match the \emph{class histogram} of $I_A$ (\texttt{same\_classes=True}), otherwise draw from the complement. This gives fine control of content resonance between $A$ and $B$.

Fixing $(I_A,I_B)$ and augmentations $(\text{aug}_A,\text{aug}_{A'},\text{aug}_B)$,
\[
\theta_A \xrightarrow{~B~} \theta_{AB}
\qquad\text{vs.}\qquad
\theta_{A'} \xrightarrow{~B~} \theta_{A'B},
\]
where $A$ and $A'$ differ only by augmentation on the \emph{same} $I_A$. We compute $D_1$ and $D_2$ on a held-out probe and report $\DeltaBF$.

In the \emph{no break} condition $B$ uses the optimizer state carried from $A$ (e.g., momentum buffers).
In the \emph{break} condition we re-initialize the optimizer immediately before $B$ (same LR/momentum), which implements the observable-level causal break discussed in \S\ref{sec:break}.

\subsection{Probe set and observable distances}
\label{sec:probe_metrics}
For each (dataset, model) we fix a probe of $N{=}2000$ held-out examples (Imagenette: validation; CIFAR-100: test). Given parameters $\theta$, we compute softmax probabilities on the probe to form $P(\theta)\in\mathbb{R}^{N\times C}$. As in \S\ref{sec:divs}, we aggregate per-example divergences to obtain $D_1$ (after the first step) and $D_2$ (after the second), and report $\DeltaBF$; $\TV$ is our primary metric, with $\JS$ and $\Hell$ for robustness checks.

We sweep four substantive regimes plus a negative control, each specifying (k,$\text{LR}$, $\text{momentum}$, $\text{aug}_A$, $\text{aug}_{A'}$, $\text{aug}_B$, $\texttt{overlap}$, $\texttt{same\_classes}$), as shown in Table~\ref{tab:regimes}.

To contextualize $\DeltaBF$, we log four diagnostics once per seed and regime (to bound cost):
\begin{enumerate}
\item \textbf{Non-commutativity ($AB\neq BA$):} for $k\in\{1,\dots,6\}$ we compare endpoints $\theta_{AB}^{(k)}$ vs.\ $\theta_{BA}^{(k)}$ via $\TV\!\big(P(\theta_{AB}^{(k)}),P(\theta_{BA}^{(k)})\big)$ on a $512$-example probe subset.
\item \textbf{Momentum alignment (no break only):} $\cos\!\big(\nabla_{\theta}\ell_B,~m\big)$ between the $B$-gradient and the optimizer’s momentum buffer just before the first $B$ step.
\item \textbf{Penultimate feature drift:} linear CKA on the input to the final linear layer for $(A,A')$ and $(AB,A'B)$ using the $512$-example probe subset.
\item \textbf{Function-space trajectory:} PCA of $\{P(\theta_A),P(\theta_{AB}),P(\theta_{A'}),P(\theta_{A'B})\}$; we plot branch centroids ($A{\to}AB$ vs.\ $A'{\to}A'B$).
\end{enumerate}

\subsection{Repeats, seeds, precision targets}
\label{sec:repeats}
For each (dataset, model, regime, break flag) and each seed $s\in\{0,1,2,3,4\}$, we run $R{=}128$ independent repeats (fresh $(I_A,I_B)$ per repeat). To control runtime while keeping tight confidence intervals, after at least $64$ repeats we check every $32$ repeats whether the nominal half-width of the 95\% CI for $\overline{\DeltaBF}$ (normal approximation) is $\le 2{\times}10^{-4}$ and stop early if so. We then aggregate per-configuration means with 2000-sample nonparametric bootstrap CIs (on per-repeat $\DeltaBF$). For completeness we also perform TOST with equivalence margin $\varepsilon{=}10^{-3}$. All micro-experiments use SGD; gradients are clipped at $1.0$; a NaN guard and a fallback LR scaling ($\times 0.5$ on retry) handle rare numerical issues. We freeze BN running stats during micro-steps, and run inference in channels-last format. The orchestration writes per-seed logs and a post-processing script emits a \texttt{summary} with mean $\DeltaBF$ and bootstrap CIs, alongside diagnostic figures\footnote{These artifacts are the basis for \S\ref{sec:results}}.

\begin{figure}[h]
  \centering
  \includegraphics[width=.40\linewidth]{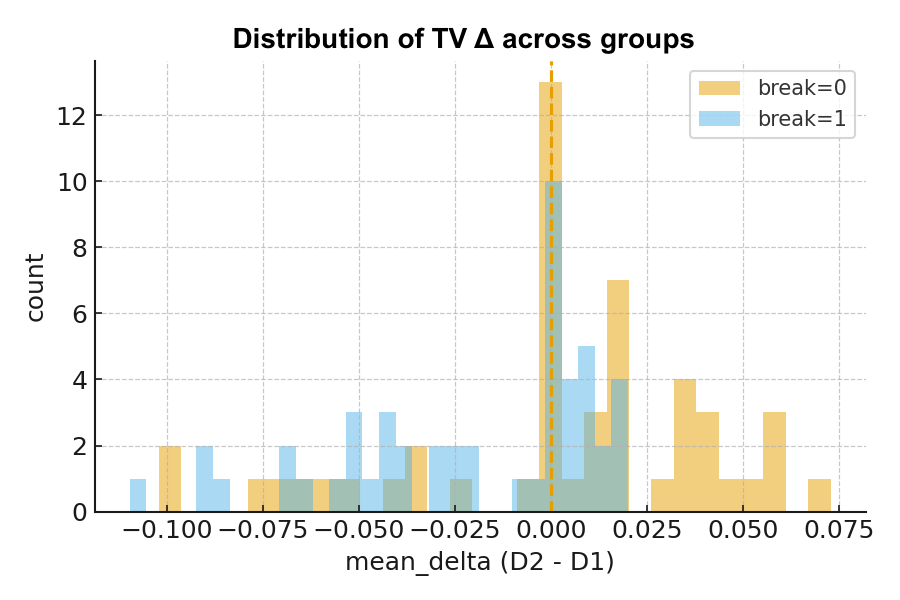}
  \caption{\textbf{Distribution of TV mean $\DeltaBF$ across all setups.}
  Histogram for \emph{no break} and \emph{break} conditions. The mass shifts toward attenuation under a causal break.}
  \label{fig:tv-dist}
\end{figure}

\begin{figure}[h]
  \centering
  
  \includegraphics[width=.40\linewidth]{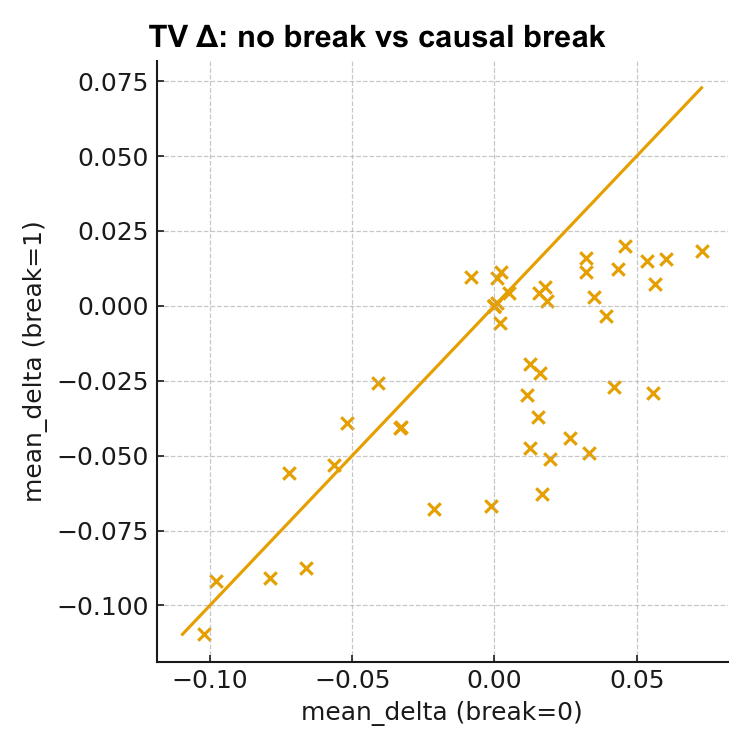}
  \caption{\textbf{TV mean effect: \emph{no break} vs.\ \emph{break}.}
  Each point is one \{\emph{dataset, model, regime, base\_stage}\}.
  Many points lie below the diagonal and $28\%$ cross quadrants (sign flips), indicating that optimizer carryover causally drives amplification.}
  \label{fig:break-scatter}
\end{figure}

\section{RESULTS}\label{sec:results}

For each configuration we report the back–flow statistic
$\DeltaBF \coloneqq \E[D_2]-\E[D_1]$
on predictive distributions over a fixed probe set ($\TV$ primary; $\JS$/$\Hell$ secondary). Intuitively, $\DeltaBF>0$ means the second instrument $B$ \emph{amplifies} the functional discrepancy induced by $A$ vs.~$A'$, whereas $\DeltaBF<0$ indicates \emph{attenuation}.
Unless stated otherwise, we show nonparametric bootstrap 95\% CIs.

Across $50$ unique \{\emph{dataset, model, regime, base\_stage}\} setups and two optimizer conditions (\emph{no break} vs.\ \emph{causal break}), we obtain $100$ TV groups.
Significance is widespread effects with narrow CIs:
with \emph{no break}, $35$ positive and $12$ negative groups have CIs excluding $0$ (only $3$ non-significant);
with a \emph{causal break}, $22$ positive and $25$ negative are significant (again $3$ non-significant).
Every unique setup is significant in at least one condition and $44/50$ are significant in \emph{both}.
Agreement across distances is strong:
TV$\wedge$Hellinger is significant in $93/100$ groups and TV$\wedge$JS in $80/100$;\footnote{TV-only significant and non-TV-only significant are rare: $0/100$ and $2/100$, respectively.}
see Fig.~\ref{fig:tv-dist}.

Introducing a \emph{causal break} (reinitializing optimizer state before $B$) systematically changes both sign and magnitude of $\DeltaBF$.
A striking $14/50$ ($28\%$) of setups \emph{flip sign} between conditions
(Fig.~\ref{fig:break-scatter}), with large absolute changes concentrated in high-momentum, high-overlap regimes (Table~\ref{tab:signflips_both})\footnote{Per-setup 95\% CIs are reported in the appendix}. Thus, \textbf{optimizer carryover is causal}: the break flips signs in 28\% of setups

\begin{table*}[h]
\caption{\textbf{Representative cases across both datasets (TV).} Rows 1–8 showcase sign flips; bottom rows give a strong attenuation baseline per dataset.}
  \label{tab:signflips_both}
  \centering
  \small
  \setlength{\tabcolsep}{5.3pt}
  \begin{tabular}{l l l l r r r}
    \toprule
    Dataset & Model & Regime & Stage & $\Delta_{\text{no}}$ & $\Delta_{\text{br}}$ & $\Delta_{\text{br}}{-}\Delta_{\text{no}}$ \\
    \midrule
    CIFAR\text{-}100 & ViT\text{-}B/16    & resonant\_strong & early & \phantom{$-$}0.0557 & $-0.0291$ & $-0.0847$ \\
    CIFAR\text{-}100 & VGG\text{-}11      & resonant\_strong & early & \phantom{$-$}0.0331 & $-0.0492$ & $-0.0823$ \\
    CIFAR\text{-}100 & ViT\text{-}B/16    & orthogonal       & early & \phantom{$-$}0.0168 & $-0.0629$ & $-0.0797$ \\
    CIFAR\text{-}100 & ViT\text{-}B/16    & resonant\_mid    & early & \phantom{$-$}0.0266 & $-0.0442$ & $-0.0708$ \\
    CIFAR\text{-}100 & MobileNetV2        & orthogonal       & early & \phantom{$-$}0.0193 & $-0.0512$ & $-0.0706$ \\
    \midrule
    Imagenette       & ViT\text{-}B/16    & orthogonal       & early & \phantom{$-$}0.0393 & $-0.0032$ & $-0.0426$ \\
    Imagenette       & ResNet\text{-}18   & standard         & early & \phantom{$-$}0.0161 & $-0.0225$ & $-0.0386$ \\
    Imagenette       & VGG\text{-}11      & standard         & early & \phantom{$-$}0.0021 & $-0.0059$ & $-0.0080$ \\
    \midrule
    CIFAR\text{-}100 & ResNet\text{-}18   & resonant\_mid    & early & $-0.1022$ & $-0.1096$ & $-0.0075$ \\
    Imagenette       & ResNet\text{-}18   & resonant\_mid    & early & $-0.0978$ & $-0.0918$ & \phantom{$-$}+0.0059 \\
    \bottomrule
  \end{tabular}
  
\end{table*}

\begin{figure}[h]
  \centering
  \includegraphics[width=.49\linewidth]{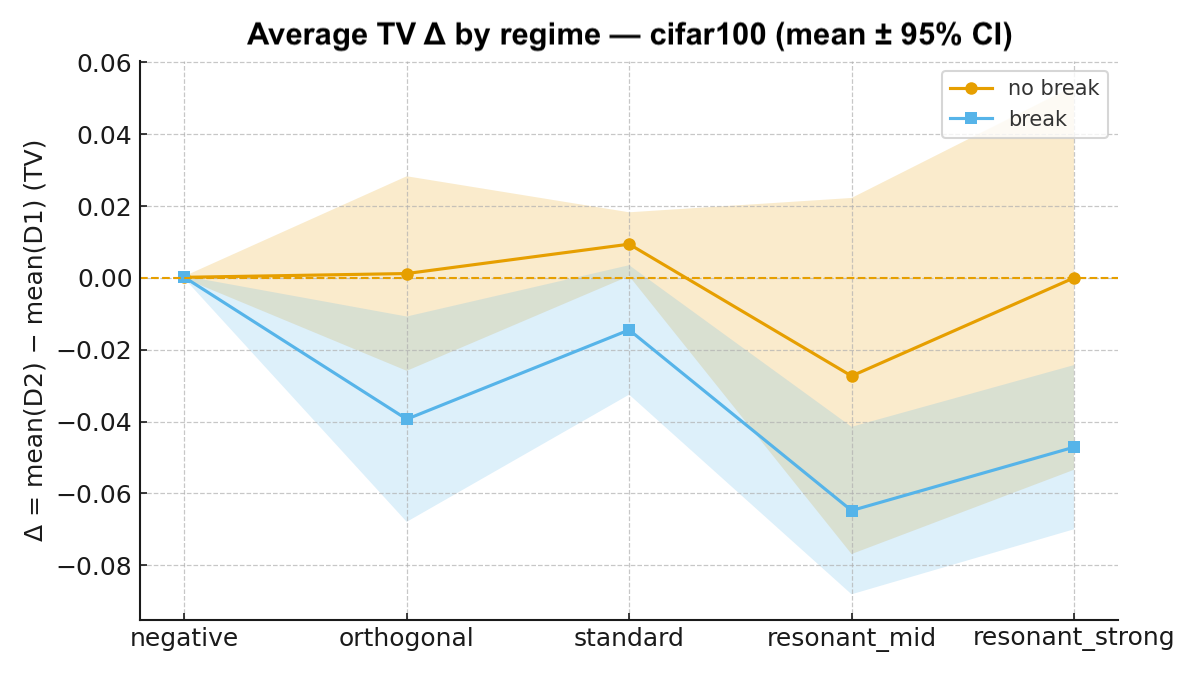}
  \includegraphics[width=.49\linewidth]{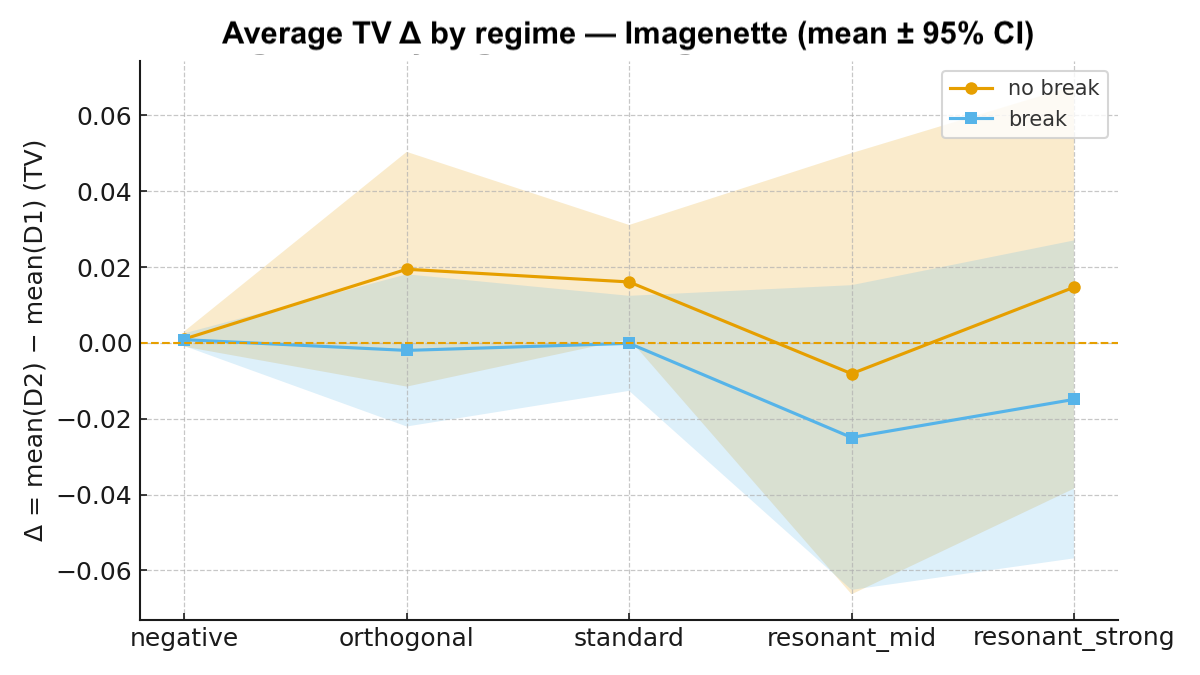}
  \caption{\textbf{Average TV effect by regime and optimizer condition.}
  Points are means across datasets and models; ribbons show bootstrap 95\% CIs.
  High-momentum/high-overlap regimes amplify without a break and attenuate with a break.}
  \label{fig:tv-regime}
\end{figure}

Averaging TV across datasets and architectures reveals clear regime effects (Fig.~\ref{fig:tv-regime}).
Without a break, \texttt{standard}, \texttt{orthogonal}, and \texttt{resonant strong} tend to \emph{amplify} ($\DeltaBF{>}0$); with a break they shift toward \emph{attenuation} ($\DeltaBF{<}0$).
The \texttt{negative} control is near zero.
Numerically (means across all dataset×model):
\texttt{standard} $+0.0127 \to -0.00729$,
\texttt{orthogonal} $+0.0103 \to -0.0206$,
\texttt{resonant strong} $+0.00734 \to -0.0310$,
\texttt{resonant mid} $-0.0177 \to -0.0448$,
\texttt{negative} $\approx 6{\times}10^{-4} \to 5{\times}10^{-4}$ (medians $\approx 10^{-7}$). Hence, we observe \textbf{regime-level trends}: momentum and overlap amplify; break attenuates.

\begin{figure}[h]
  \centering
  \includegraphics[width=.4\linewidth]{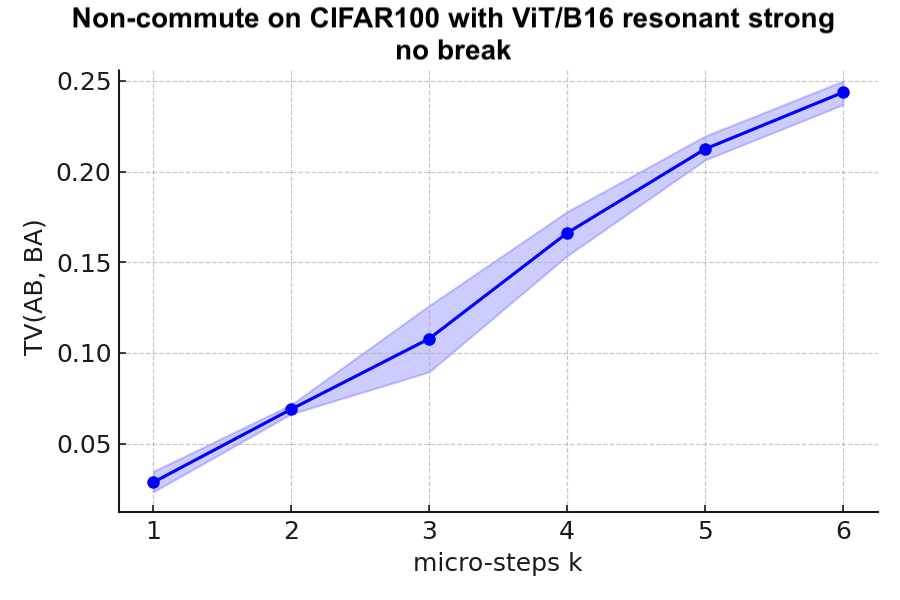}\hfill
  \includegraphics[width=.4\linewidth]{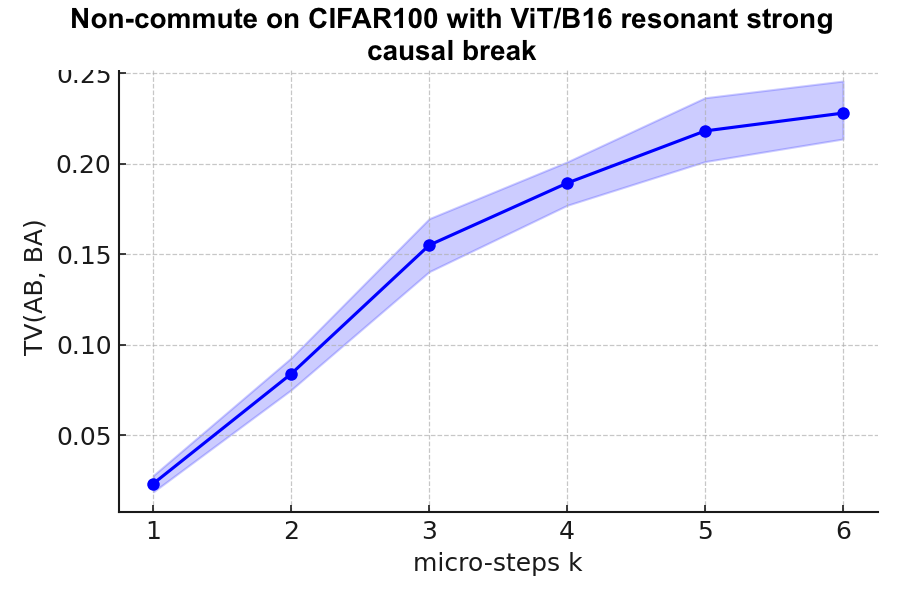}
  \caption{\textbf{Non-commute curves.}
  TV$(AB,BA)$ vs.\ micro-steps $k$ for \texttt{CIFAR-100/ViT-B/16/resonant strong}. Left: \emph{no break}. Right: \emph{break}.}
  \label{fig:noncomm}
\end{figure}

\begin{figure}[h]
  \centering
  \includegraphics[width=.4\linewidth]{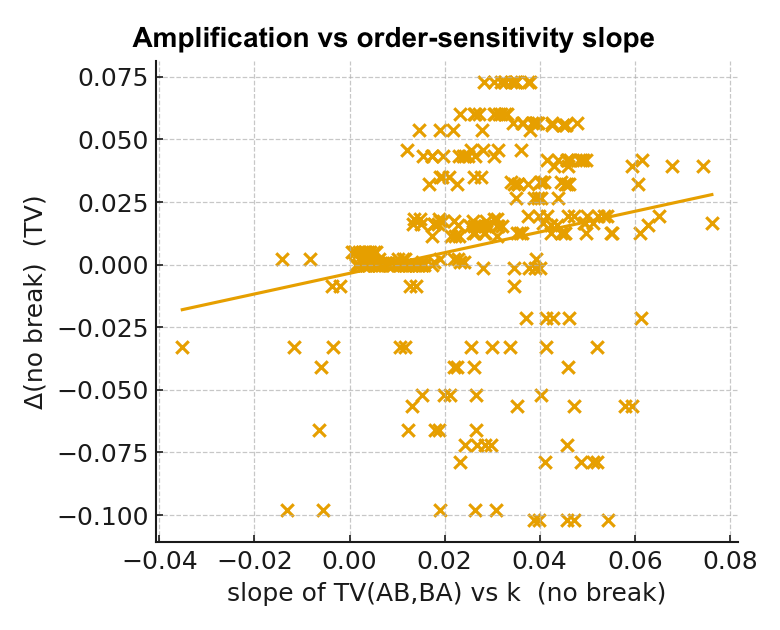}
  \includegraphics[width=.4\linewidth]{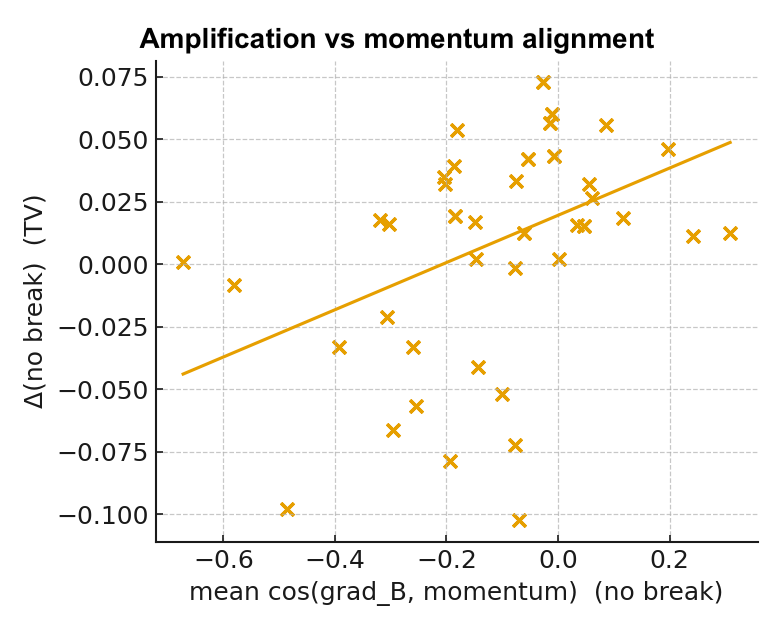}
  \caption{\textbf{Amplification vs. order sensitivity (left).} Configurations with a steeper \(\mathrm{TV}(AB,BA)\) slope (no break) exhibit larger \(\Delta\). \textbf{Amplification vs. momentum alignment (right).} Each point corresponds to a configuration. Greater pre-\(B\) alignment predicts larger \(\Delta\). Least-squares fits are shown.}

  \label{fig:delta-vs-noncomm}
\end{figure}

\begin{figure}[h]
  \centering
  \includegraphics[width=.5\linewidth]{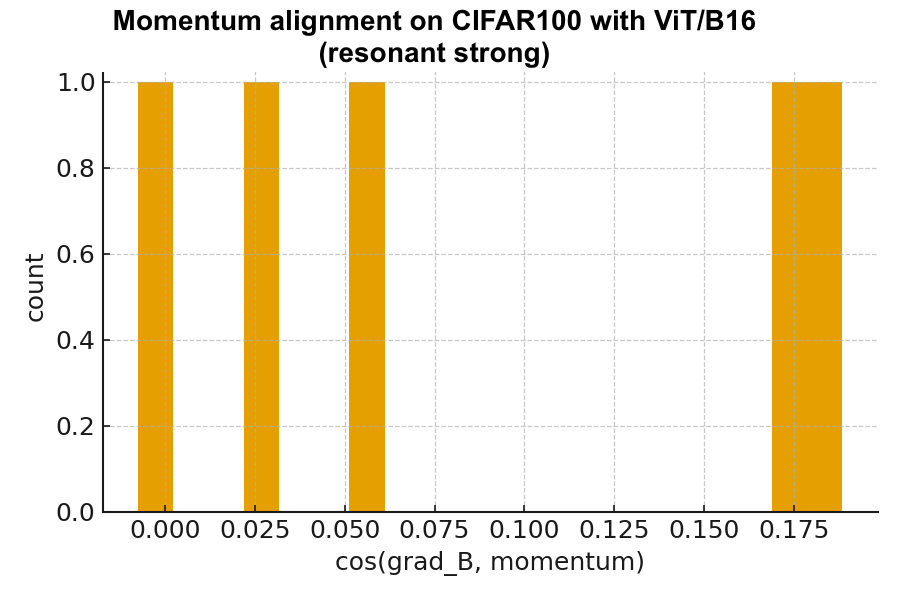}
  \caption{\textbf{Momentum alignment (no break).}
  Histogram of $\cos(\nabla_B, m)$; mass $>0$ indicates optimizer carryover aligns with the upcoming $B$ update.}
  \label{fig:mom}
\end{figure}

\begin{figure}[h]
  \centering
  \includegraphics[width=.4\linewidth]{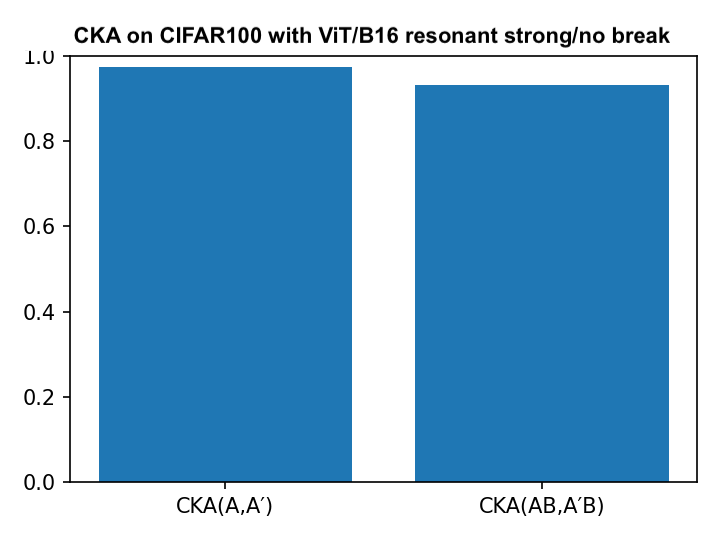}\hfill
  \includegraphics[width=.4\linewidth]{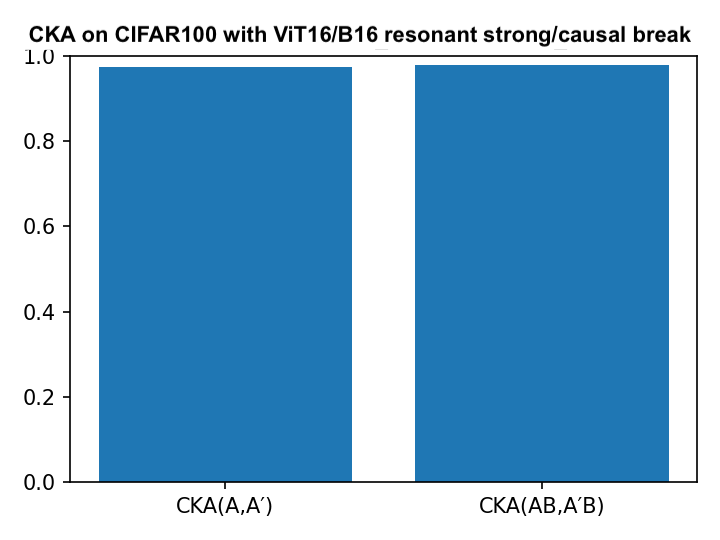}
  \includegraphics[width=.4\linewidth]{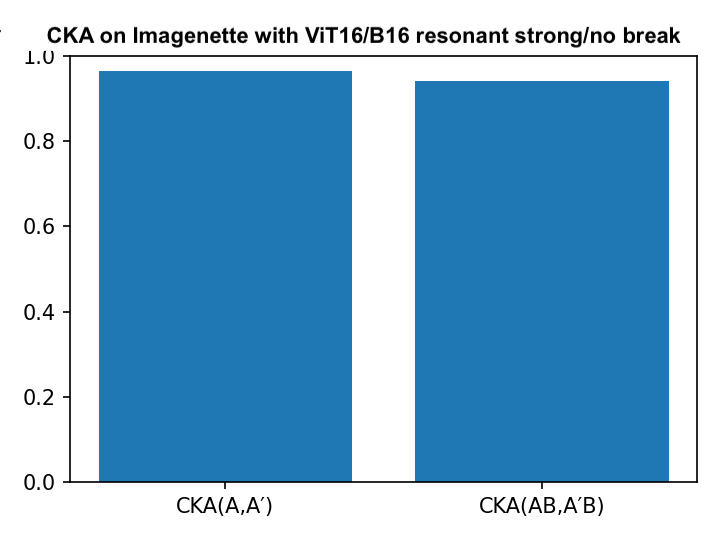}\hfill
  \includegraphics[width=.4\linewidth]{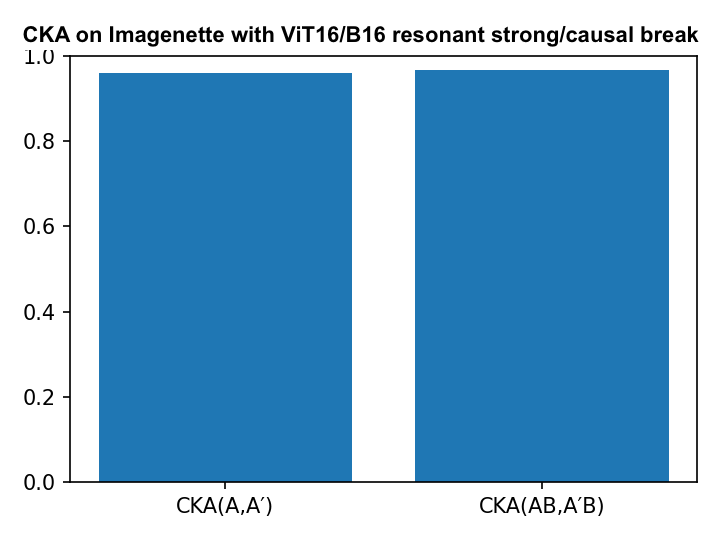}
  \caption{\textbf{Penultimate-layer CKA.}
  Left: \emph{no break}. Right: \emph{break}. Top: CIFAR-100. Bottom: Imagenette. Break increases CKA$(AB,A'B)$ relative to CKA$(A,A')$.}
  \label{fig:cka}
\end{figure}

\begin{figure}[h]
  \centering
  \includegraphics[width=.4\linewidth]{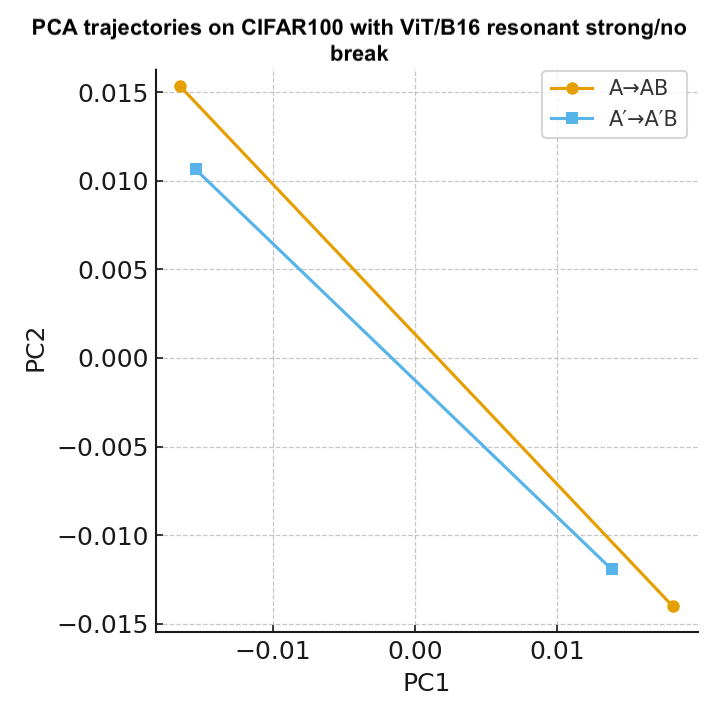}\hfill
  \includegraphics[width=.4\linewidth]{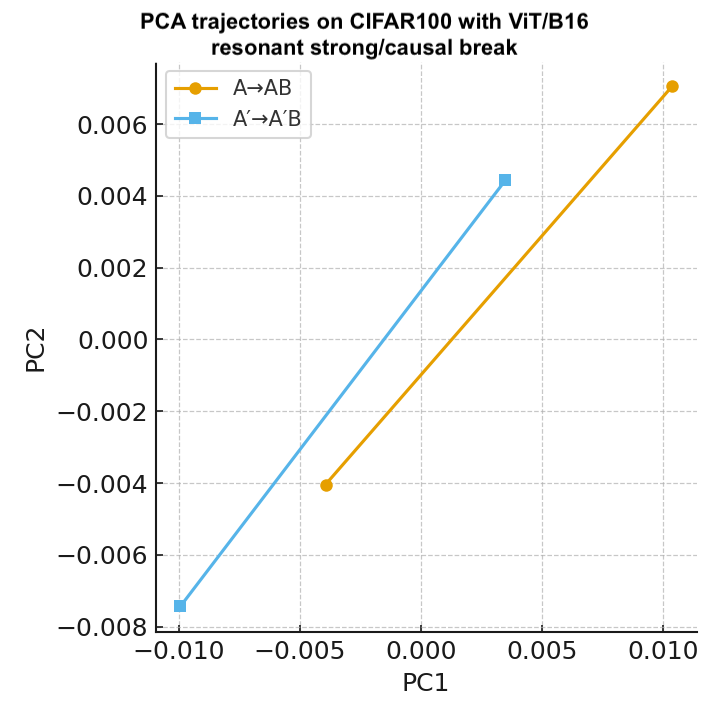}
  \caption{\textbf{Function-space trajectories.}
  Displacements and branch separation are larger without a break (left); a break shortens and aligns paths (right).}
  \label{fig:traj}
\end{figure}

For the setups in Fig.~\ref{fig:noncomm}, the non-commute curve TV$(AB,BA)$ increases with the number of micro-steps $k$ , evidencing order sensitivity even at short horizons. The slope is typically steeper without a break and flattens with a break, consistent with optimizer carryover as a driver of non-commutativity. Configurations whose non-commute curve grows faster with micro-step depth also exhibit larger \(\Delta\). The slope of \(\mathrm{TV}(AB,BA)\) vs.\ \(k\) (no break) correlates with \(\Delta\): Pearson \(r=0.184\) (p \(=1.38\times10^{-3}\)) and Spearman \(\rho=0.329\) (p \(=5.45\times10^{-9}\)) across configurations (Fig.~\ref{fig:delta-vs-noncomm} (left)). This links the back-flow witness to order-dependence.

We directly log the cosine between the $B$-gradient and the momentum buffer just before the first $B$ step.
Without a break, the distribution concentrates above zero (Fig.~\ref{fig:mom}), diagnosing \emph{alignment} that explains amplification ($\DeltaBF{>}0$).
With a break, alignment vanishes by construction and $\DeltaBF$ attenuates or flips. Beyond the qualitative histogram, \(\Delta\) scales with pre-\(B\) momentum alignment: across configurations (dataset\(\times\)model\(\times\)regime; no-break), we find Pearson \(r=0.409\) (p \(=4.09\times10^{-11}\)) and Spearman \(\rho=0.454\) (p \(=1.37\times10^{-13}\)) between \(\Delta\) and the mean \(\cos(\nabla_B,m)\) (Fig.~\ref{fig:delta-vs-noncomm} (right)). This dose–response matches the linearized prediction in \S\ref{sec:perturb}. After a causal break, alignment is null by construction and the amplification collapses. To probe feature-space consequences, we compare penultimate-layer linear CKA for $(A,A')$ vs.\ $(AB,A'B)$.
Under a break, CKA$(AB,A'B)$ typically exceeds CKA$(A,A')$ (realignment/attenuation);
without a break, the ordering often reverses (divergence/amplification), cf.\ Fig.~\ref{fig:cka}.
Function-space PCA trajectories show the same pattern: longer $A{\to}AB$ displacements and wider branch separation without a break; convergence or crossing with a break (Fig.~\ref{fig:traj}).

The \texttt{negative} regime behaves as a control: means $\approx (5\text{--}6)\!\times\!10^{-4}$ and medians $\approx 10^{-7}$ under both conditions.
We observe multiple narrow-CI cases (95\% CI half-width $\le 2\!\times\!10^{-4}$ in nine per condition), supporting the stability of the estimates.
TOST with margin $\varepsilon=10^{-3}$ classifies many break-condition effects as practically null, while rejecting nullity for no-break amplification; see Appendix for per-setup statistics. 

As a falsification check, placebo runs with \(A{=}A'\) (identical augmentations) yield \(\Delta\) statistically indistinguishable from zero (95\% CIs spanning \(0\)); likewise, no-break runs with \(\mu{=}0\) collapse to the break baseline, consistent with optimizer buffers mediating back-flow. Regarding the limitations of this work we restrict to short two-step interventions, image classification benchmarks, and standard vision backbones; we study the existence and controllability of memory rather than downstream benefits. Our diagnostic complements, not replaces, SDE/Markov analyses, and it inherits the usual caveats of probe design (choice of $B$, probe set, metric).

\section{Conclusions}

The memory we detect is \emph{intentional in mechanism}---optimizer state is carried  across steps by design---but \emph{unintended in consequence}: it induces operational non-Markovianity, whereby the effect of a fixed second intervention $B$ depends on the immediate history $A$. Whether such memory is helpful or harmful depends on the alignment between training order and target generalization. Our witness $\Delta_{\mathrm{BF}}$ offers a direct diagnostic: it quantifies when optimizer memory amplifies prior updates (as in ``resonant'' schedules), and when it should be collapsed with a causal break before switching regimes. We identify optimizer state as a \emph{causal driver} of observable back-flow. It systematically amplifies the influence of $A$ on subsequent $B$, induces order 
sensitivity ($AB \neq BA$), and can be neutralized by a simple reset of optimizer buffers. This establishes a practical intervention that links mechanism and measurement. 
The phenomenon is robust: we observe consistent effects across datasets (CIFAR-100, Imagenette), architectures (SmallCNN, ResNet-18, VGG-11, MobileNetV2,  ViT-B/16), divergences (TV, JS, H), and training regimes. These results demonstrate that optimizer-induced memory is a pervasive and measurable feature of modern training dynamics, and that back-flow provides a principled, operational handle for studying and controlling it.

\bibliographystyle{unsrt}  
\bibliography{references}  
\appendix

\section{PROOFS}
\textbf{Proposition 3.1}
Let $\mathcal{Z}$ be a (standard Borel) observation space for the probe observable, and let $\mathcal{P}(\mathcal{Z})$ denote the set of probability measures on $\mathcal{Z}$. 
Fix a family of first-step instruments $\mathcal{I}$ and a common second-step instrument $B$.
For each $I\in\mathcal{I}$, let $P_I\in\mathcal{P}(\mathcal{Z})$ be the (pre-$B$) predictive distribution on the probe, and let $Q_I\in\mathcal{P}(\mathcal{Z})$ be the (post-$B$) predictive distribution.
Assume the two-time process is \emph{operationally Markov at the observable for $B$}, namely:
there exists a Markov kernel (classical channel) $\Lambda_B:\mathcal{P}(\mathcal{Z})\to\mathcal{P}(\mathcal{Z})$ such that
\[
Q_I=\Lambda_B(P_I)\qquad\text{for all } I\in\mathcal{I}.
\]
Let $D:\mathcal{P}(\mathcal{Z})\times\mathcal{P}(\mathcal{Z})\to[0,\infty]$ be any divergence that is \emph{contractive under classical channels}, i.e., $D(\Lambda p,\Lambda q)\le D(p,q)$ for all probability measures $p,q$ and Markov kernels $\Lambda$.
Define the back-flow of distinguishability for any pair $I,I'\in\mathcal{I}$ by
\[
\Delta^{(I,B)}(I,I')\;\coloneqq\;D(Q_I,Q_{I'})-D(P_I,P_{I'})\, .
\]
Then $\Delta^{(I,B)}(I,I')\le 0$ for all $I,I'\in\mathcal{I}$.

\textbf{Proof}
By operational Markovianity at the observable for $B$, there exists a \emph{fixed} Markov kernel $\Lambda_B$ such that $Q_I=\Lambda_B(P_I)$ and $Q_{I'}=\Lambda_B(P_{I'})$ for all instruments $I,I'$. 
By contractivity of $D$ under classical channels (data-processing inequality),
\[
D(Q_I,Q_{I'}) \;=\; D\bigl(\Lambda_B(P_I),\,\Lambda_B(P_{I'})\bigr)\;\le\; D(P_I,P_{I'})\, .
\]
Rearranging gives $D(Q_I,Q_{I'})-D(P_I,P_{I'})\le 0$, i.e., $\Delta^{(I,B)}(I,I')\le 0$.
Since $I,I'$ were arbitrary, the claim holds for all pairs.

\paragraph{Remarks.}
(i) In the discrete case, $\Lambda_B$ is a stochastic matrix acting on probability vectors, and the proof reduces to $D(\Lambda_B p,\Lambda_B q)\le D(p,q)$.
(ii) The assumption on $D$ holds for standard contractive divergences (e.g., total variation, Hellinger, Jensen--Shannon, and more generally Csisz{\'a}r $f$-divergences under mild regularity).
(iii) Equality $\Delta^{(I,B)}=0$ occurs, e.g., when $\Lambda_B$ is \emph{sufficient} for the pair $(P_I,P_{I'})$ in the sense that it preserves their statistical distance as measured by $D$.

\subsection{Proof of Proposition 3.2}

\textbf{Proposition 3.2}
Let $D$ be any divergence on probability measures that satisfies data processing: 
$D(\Lambda p,\Lambda q)\le D(p,q)$ for all Markov kernels $\Lambda$ and all $p,q$.
Assume the operational Markov condition \eqref{eq:omc} for the second-step observable $B$, i.e., there exists a (fixed) classical channel $\Lambda_B$ such that 
\begin{equation}\label{eq:omc-restate}
\Phi_2^{(I,B)}=\Lambda_B\,\Phi_1^{(I)}\qquad\text{for every first-step instrument }I.
\end{equation}
For any two first-step instruments $A,A'$, define the back-flow of distinguishability
\[
\Delta^{(I,B)} \;\coloneqq\;
D\!\bigl(\Phi_2^{(A,B)},\Phi_2^{(A',B)}\bigr)
\;-\;
D\!\bigl(\Phi_1^{(A)},\Phi_1^{(A')}\bigr).
\]
Then $\Delta^{(I,B)}\le 0$.

\textbf{Proof}
Fix any two first-step instruments $A,A'$. Set $P=\Phi_1^{(A)}$ and $Q=\Phi_1^{(A')}$.
By \eqref{eq:omc-restate}, 
\[
\Phi_2^{(A,B)}=\Lambda_B P
\qquad\text{and}\qquad
\Phi_2^{(A',B)}=\Lambda_B Q.
\]
Applying data processing for $D$ with the channel $\Lambda_B$ yields
\[
D\!\bigl(\Phi_2^{(A,B)},\Phi_2^{(A',B)}\bigr)
= D(\Lambda_B P,\Lambda_B Q)
\;\le\; D(P,Q)
= D\!\bigl(\Phi_1^{(A)},\Phi_1^{(A')}\bigr).
\]
Rearranging gives $\Delta^{(I,B)}\le 0$, as claimed.

\paragraph{Remarks.}
(1) The argument is agnostic to the sample space (finite or general Borel) and to the particular choice of $D$, provided $D$ obeys data processing (e.g., total variation, Hellinger, Jensen--Shannon, and Csiszár $f$-divergences under standard regularity). 
(2) If you use the metric $\sqrt{\mathrm{JS}}$, data processing holds and the same conclusion follows verbatim. 
(3) If you later introduce an \emph{approximate} OMC, $\|\Phi_2^{(I,B)} - \Lambda_B \Phi_1^{(I)}\|_D \le \eta_I$ for all $I$ and a triangle-inequality divergence $\|\cdot\|_D$, then 
\[
\Delta^{(I,B)} \le \eta_A+\eta_{A'},
\]
by a two-sided triangle inequality combined with contractivity of $D$; this yields a stable, quantitative relaxation of Proposition~\ref{prop:nobackflow}.

\textbf{Proof of Theorem 3.4}
Let $\mathcal Q,\mathcal O$ be standard Borel spaces. 
Let $\mathcal O:\mathcal Q\rightsquigarrow\mathcal O$ be the probe observable (a Markov kernel) and let $\mathcal K_B:\mathcal Q\rightsquigarrow\mathcal Q$ be the second-step dynamics for instrument $B$.
For a first-step instrument $I$, write $\pi_1^{(I)}\in\mathcal P(\mathcal Q)$ for the latent state after step~1 and $\Phi_1^{(I)}\coloneqq \pi_1^{(I)}\mathcal O\in\mathcal P(\mathcal O)$ for the corresponding predictive (pre-$B$) observable law. 
Assume:
\begin{enumerate}
\item[\textnormal{(Break)}] \emph{Causal break} $\mathcal B$ is applied between step~1 and $B$, so that the post-break latent state depends on $I$ only via $\Phi_1^{(I)}$, i.e., there exists a Markov kernel $\mathcal R:\mathcal O\rightsquigarrow\mathcal Q$ with
\[
\widetilde \pi_1^{(I)} \;=\; \Phi_1^{(I)}\,\mathcal R \qquad\text{for all } I.
\]
\item[\textnormal{(Suff)}] \emph{Sufficiency of $\mathcal O$ for $B$ on $\mathcal Q$}: there exists a Markov kernel $\Lambda_B:\mathcal O\rightsquigarrow\mathcal O$ such that
\begin{equation}\label{eq:suff}
\mu\,\mathcal R\,\mathcal K_B\,\mathcal O \;=\; \mu\,\Lambda_B
\qquad\text{for all }\mu\in\mathcal P(\mathcal O).
\end{equation}
(Equivalently, for all $\pi\in\mathcal P(\mathcal Q)$ one has $\pi \mathcal K_B \mathcal O = (\pi\mathcal O)\Lambda_B$.)
\end{enumerate}
Then for all first-step instruments $I$,
\[
\Phi_2^{(I,B)} \;=\; \Lambda_B\,[\,\Phi_1^{(I)}\,].
\]
Consequently, for any divergence $D$ that is contractive under classical channels (data processing), the back-flow satisfies $\Delta^{(I,B)}\le 0$.

\textbf{Proof}
By the causal break, the latent state just before applying $B$ is $\widetilde \pi_1^{(I)}=\Phi_1^{(I)}\mathcal R$. 
Pushing it forward through $B$ and then observing via $\mathcal O$ gives
\[
\Phi_2^{(I,B)} \;=\; \widetilde \pi_1^{(I)}\,\mathcal K_B\,\mathcal O 
\;=\; \bigl(\Phi_1^{(I)}\,\mathcal R\bigr)\,\mathcal K_B\,\mathcal O 
\;=\; \Phi_1^{(I)}\,(\mathcal R\,\mathcal K_B\,\mathcal O).
\]
By sufficiency \eqref{eq:suff}, the composite $\mathcal R\,\mathcal K_B\,\mathcal O$ factors through a \emph{fixed} observable channel $\Lambda_B:\mathcal O\rightsquigarrow\mathcal O$ independent of $I$, hence
\[
\Phi_2^{(I,B)} \;=\; \Phi_1^{(I)}\,\Lambda_B \;=\; \Lambda_B[\,\Phi_1^{(I)}\,].
\]
This proves the first claim.

For the back-flow bound, let $A,A'$ be any two first-step instruments and set $P=\Phi_1^{(A)}$, $Q=\Phi_1^{(A')}$.
Then
\[
\Phi_2^{(A,B)}=\Lambda_B P,\qquad \Phi_2^{(A',B)}=\Lambda_B Q.
\]
By data processing (contractivity) of $D$ under classical channels,
\[
D\!\bigl(\Phi_2^{(A,B)},\Phi_2^{(A',B)}\bigr) 
\;=\; D(\Lambda_B P,\Lambda_B Q) 
\;\le\; D(P,Q)
\;=\; D\!\bigl(\Phi_1^{(A)},\Phi_1^{(A')}\bigr).
\]
Rearranging yields $\Delta^{(I,B)}\le 0$.

\paragraph{Notes on assumptions.}
\begin{itemize}
\item The \emph{causal break} is encoded by $\mathcal R:\mathcal O\rightsquigarrow\mathcal Q$; it re-prepares the latent state using only the observed law from step~1, thus erasing any dependence on the pre-history beyond $\Phi_1^{(I)}$.
\item The \emph{sufficiency} condition is the standard Blackwell–Sherman–Stein notion: the pair $(\mathcal K_B,\mathcal O)$ \emph{garbles} through the statistic $\mathcal O$, i.e.\ there exists a channel $\Lambda_B$ on observables such that for all priors $\pi$ on $\mathcal Q$, $(\pi\mathcal O)\Lambda_B=\pi\mathcal K_B\mathcal O$. This yields \eqref{eq:suff} globally on $\mathcal P(\mathcal O)$ (no extension argument is needed).
\item In discrete settings, $\mathcal R,\mathcal K_B,\mathcal O,\Lambda_B$ are stochastic matrices and the proof reduces to associativity of matrix multiplication.
\end{itemize}

\section{ADDITIONAL TABLES}

\begin{longtable}{l l l l c c c c}
\caption{\textbf{Per-group significance by metric.} A checkmark means the 95\% CI for $\Delta$ excludes $0$. Break: \texttt{no} vs \texttt{br}.}\label{tab:per-group-metric-sig}\\
\toprule
Dataset & Model & Regime & Stage & Break & TV & JS & Hell \\
\midrule
\endfirsthead
\toprule
Dataset & Model & Regime & Stage & Break & TV & JS & Hell \\
\midrule
\endhead
\midrule
\multicolumn{8}{r}{\emph{continued on next page}}\\
\midrule
\endfoot
\bottomrule
\endlastfoot
cifar100 & mobilenetv2 & negative & early & br & \checkmark & \checkmark & \checkmark \\
cifar100 & mobilenetv2 & negative & early & no & \checkmark & \checkmark & \checkmark \\
cifar100 & mobilenetv2 & orthogonal & early & br & \checkmark & \checkmark & \checkmark \\
cifar100 & mobilenetv2 & orthogonal & early & no & \checkmark & \checkmark & \checkmark \\
cifar100 & mobilenetv2 & resonant mid & early & br & \checkmark & \checkmark & \checkmark \\
cifar100 & mobilenetv2 & resonant mid & early & no & \checkmark & \checkmark & \checkmark \\
cifar100 & mobilenetv2 & resonant strong & early & br & \checkmark & \checkmark & \checkmark \\
cifar100 & mobilenetv2 & resonant strong & early & no & \checkmark & \checkmark & \checkmark \\
cifar100 & mobilenetv2 & standard & early & br & \checkmark & \checkmark & \checkmark \\
cifar100 & mobilenetv2 & standard & early & no & \checkmark & \checkmark & \checkmark \\
cifar100 & resnet18 & negative & early & br & \checkmark & \checkmark & \checkmark \\
cifar100 & resnet18 & negative & early & no & \checkmark & \checkmark & \checkmark \\
cifar100 & resnet18 & orthogonal & early & br & \checkmark & \checkmark & \checkmark \\
cifar100 & resnet18 & orthogonal & early & no & \checkmark & \checkmark & \checkmark \\
cifar100 & resnet18 & resonant mid & early & br & \checkmark & \checkmark & \checkmark \\
cifar100 & resnet18 & resonant mid & early & no & \checkmark & \checkmark & \checkmark \\
cifar100 & resnet18 & resonant strong & early & br & \checkmark & \checkmark & \checkmark \\
cifar100 & resnet18 & resonant strong & early & no & \checkmark & \checkmark & \checkmark \\
cifar100 & resnet18 & standard & early & br & -- & \checkmark & -- \\
cifar100 & resnet18 & standard & early & no & \checkmark & \checkmark & \checkmark \\
cifar100 & smallcnn & negative & early & br & \checkmark & -- & \checkmark \\
cifar100 & smallcnn & negative & early & no & \checkmark & -- & \checkmark \\
cifar100 & smallcnn & orthogonal & early & br & -- & -- & -- \\
cifar100 & smallcnn & orthogonal & early & no & \checkmark & \checkmark & \checkmark \\
cifar100 & smallcnn & resonant mid & early & br & \checkmark & \checkmark & \checkmark \\
cifar100 & smallcnn & resonant mid & early & no & \checkmark & \checkmark & \checkmark \\
cifar100 & smallcnn & resonant strong & early & br & \checkmark & \checkmark & \checkmark \\
cifar100 & smallcnn & resonant strong & early & no & \checkmark & \checkmark & \checkmark \\
cifar100 & smallcnn & standard & early & br & \checkmark & \checkmark & \checkmark \\
cifar100 & smallcnn & standard & early & no & \checkmark & \checkmark & \checkmark \\
cifar100 & vgg11 & negative & early & br & \checkmark & \checkmark & \checkmark \\
cifar100 & vgg11 & negative & early & no & \checkmark & -- & \checkmark \\
cifar100 & vgg11 & orthogonal & early & br & \checkmark & \checkmark & \checkmark \\
cifar100 & vgg11 & orthogonal & early & no & \checkmark & \checkmark & \checkmark \\
cifar100 & vgg11 & resonant mid & early & br & \checkmark & \checkmark & \checkmark \\
cifar100 & vgg11 & resonant mid & early & no & -- & -- & -- \\
cifar100 & vgg11 & resonant strong & early & br & \checkmark & \checkmark & \checkmark \\
cifar100 & vgg11 & resonant strong & early & no & \checkmark & \checkmark & \checkmark \\
cifar100 & vgg11 & standard & early & br & \checkmark & \checkmark & \checkmark \\
cifar100 & vgg11 & standard & early & no & \checkmark & \checkmark & \checkmark \\
cifar100 & vit-b-16 & negative & early & br & \checkmark & -- & \checkmark \\
cifar100 & vit-b-16 & negative & early & no & \checkmark & -- & \checkmark \\
cifar100 & vit-b-16 & orthogonal & early & br & \checkmark & \checkmark & \checkmark \\
cifar100 & vit-b-16 & orthogonal & early & no & \checkmark & \checkmark & \checkmark \\
cifar100 & vit-b-16 & resonant mid & early & br & \checkmark & \checkmark & \checkmark \\
cifar100 & vit-b-16 & resonant mid & early & no & \checkmark & \checkmark & \checkmark \\
cifar100 & vit-b-16 & resonant strong & early & br & \checkmark & \checkmark & \checkmark \\
cifar100 & vit-b-16 & resonant strong & early & no & \checkmark & \checkmark & \checkmark \\
cifar100 & vit-b-16 & standard & early & br & \checkmark & \checkmark & \checkmark \\
cifar100 & vit-b-16 & standard & early & no & \checkmark & \checkmark & \checkmark \\
imagenette & mobilenetv2 & negative & early & br & \checkmark & \checkmark & \checkmark \\
imagenette & mobilenetv2 & negative & early & no & \checkmark & \checkmark & \checkmark \\
imagenette & mobilenetv2 & orthogonal & early & br & \checkmark & \checkmark & \checkmark \\
imagenette & mobilenetv2 & orthogonal & early & no & \checkmark & \checkmark & \checkmark \\
imagenette & mobilenetv2 & resonant mid & early & br & \checkmark & \checkmark & \checkmark \\
imagenette & mobilenetv2 & resonant mid & early & no & \checkmark & \checkmark & \checkmark \\
imagenette & mobilenetv2 & resonant strong & early & br & \checkmark & \checkmark & \checkmark \\
imagenette & mobilenetv2 & resonant strong & early & no & \checkmark & \checkmark & \checkmark \\
imagenette & mobilenetv2 & standard & early & br & \checkmark & \checkmark & \checkmark \\
imagenette & mobilenetv2 & standard & early & no & \checkmark & \checkmark & \checkmark \\
imagenette & resnet18 & negative & early & br & \checkmark & -- & \checkmark \\
imagenette & resnet18 & negative & early & no & \checkmark & -- & \checkmark \\
imagenette & resnet18 & orthogonal & early & br & \checkmark & \checkmark & \checkmark \\
imagenette & resnet18 & orthogonal & early & no & \checkmark & \checkmark & \checkmark \\
imagenette & resnet18 & resonant mid & early & br & \checkmark & \checkmark & \checkmark \\
imagenette & resnet18 & resonant mid & early & no & \checkmark & \checkmark & \checkmark \\
imagenette & resnet18 & resonant strong & early & br & \checkmark & \checkmark & \checkmark \\
imagenette & resnet18 & resonant strong & early & no & \checkmark & \checkmark & \checkmark \\
imagenette & resnet18 & standard & early & br & \checkmark & \checkmark & \checkmark \\
imagenette & resnet18 & standard & early & no & \checkmark & \checkmark & \checkmark \\
imagenette & smallcnn & negative & early & br & \checkmark & -- & \checkmark \\
imagenette & smallcnn & negative & early & no & \checkmark & -- & \checkmark \\
imagenette & smallcnn & orthogonal & early & br & \checkmark & \checkmark & \checkmark \\
imagenette & smallcnn & orthogonal & early & no & -- & -- & -- \\
imagenette & smallcnn & resonant mid & early & br & \checkmark & \checkmark & \checkmark \\
imagenette & smallcnn & resonant mid & early & no & \checkmark & \checkmark & \checkmark \\
imagenette & smallcnn & resonant strong & early & br & \checkmark & \checkmark & \checkmark \\
imagenette & smallcnn & resonant strong & early & no & \checkmark & \checkmark & \checkmark \\
imagenette & smallcnn & standard & early & br & \checkmark & \checkmark & \checkmark \\
imagenette & smallcnn & standard & early & no & -- & -- & -- \\
imagenette & vgg11 & negative & early & br & \checkmark & -- & \checkmark \\
imagenette & vgg11 & negative & early & no & \checkmark & -- & \checkmark \\
imagenette & vgg11 & orthogonal & early & br & \checkmark & \checkmark & \checkmark \\
imagenette & vgg11 & orthogonal & early & no & \checkmark & \checkmark & \checkmark \\
imagenette & vgg11 & resonant mid & early & br & -- & \checkmark & -- \\
imagenette & vgg11 & resonant mid & early & no & \checkmark & \checkmark & \checkmark \\
imagenette & vgg11 & resonant strong & early & br & \checkmark & \checkmark & \checkmark \\
imagenette & vgg11 & resonant strong & early & no & \checkmark & \checkmark & \checkmark \\
imagenette & vgg11 & standard & early & br & \checkmark & \checkmark & \checkmark \\
imagenette & vgg11 & standard & early & no & \checkmark & \checkmark & \checkmark \\
imagenette & vit-b-16 & negative & early & br & \checkmark & -- & \checkmark \\
imagenette & vit-b-16 & negative & early & no & \checkmark & -- & \checkmark \\
imagenette & vit-b-16 & orthogonal & early & br & \checkmark & -- & \checkmark \\
imagenette & vit-b-16 & orthogonal & early & no & \checkmark & \checkmark & \checkmark \\
imagenette & vit-b-16 & resonant mid & early & br & \checkmark & \checkmark & -- \\
imagenette & vit-b-16 & resonant mid & early & no & \checkmark & \checkmark & \checkmark \\
imagenette & vit-b-16 & resonant strong & early & br & \checkmark & \checkmark & \checkmark \\
imagenette & vit-b-16 & resonant strong & early & no & \checkmark & \checkmark & \checkmark \\
imagenette & vit-b-16 & standard & early & br & \checkmark & \checkmark & \checkmark \\
imagenette & vit-b-16 & standard & early & no & \checkmark & \checkmark & \checkmark \\
\end{longtable}

\begin{table}[h]
\caption{\textbf{Metric agreement summary.} Counts over all \{dataset, model, regime, stage, break\} groups.}
  \label{tab:metric-agreement}
  \centering
  \small
  \begin{tabular}{l r}
    \toprule
    Total groups & 100 \\
    \midrule
    TV significant & 97 \\
    JS significant & 82 \\
    Hellinger significant & 94 \\
    \midrule
    TV $\wedge$ Hellinger significant & 93 / 100 \\
    TV $\wedge$ JS significant & 80 / 100 \\
    All three significant & 80 / 100 \\
    \midrule
    TV-only significant & 0 / 100 \\
    non-TV-only significant & 2 / 100 \\
    None significant & 4 / 100 \\
    \bottomrule
  \end{tabular}
  
\end{table}

For each \{dataset, model, regime, stage, break\} configuration we report the mean $\overline{\Delta}$, a nonparametric 95\% bootstrap CI, the number of repeats $n$ (pooled across seeds), the CI half-width, and TOST results for $\varepsilon=10^{-3}$; see Table~\ref{tab:per-setup-tv-fdr}.
A compact control summary for the \texttt{negative} regime (means and medians under both optimizer conditions) appears in Table~\ref{tab:negative-control}.

\begin{table}[h]
\centering
\small
\begin{tabular}{l r r r}
\toprule
Break & Mean $\overline{\Delta}$ & Median $\overline{\Delta}$ & Count \\
\midrule
no & 0.000597 & 0.000000 & 10 \\
br & 0.000516 & 0.000000 & 10 \\
\bottomrule
\end{tabular}
\caption{\textbf{Negative regime as control (TV).} Means and medians across setups under no-break and break conditions.}
\label{tab:negative-control}
\end{table}

We complement bootstrap CIs with Benjamini--Hochberg (BH) false-discovery-rate control 
applied to two-sided tests of $H_{0}:\,\mathbb{E}[\Delta] = 0$ per configuration. 
At FDR $5\%$, TV: $94/100$, Hellinger: $93/100$, and JS: $81/100$ groups remain significant; 
conclusions are unchanged. Per-setup tables report raw $p$ and BH $q$-values alongside 
$95\%$ bootstrap CIs.


\begin{landscape}
\begin{longtable}{l l l l c r c r l l}
\caption{\textbf{Per-setup statistics with FDR (TV).} Mean $\Delta$, 95\% bootstrap CI, $n$, raw two-sided $p$, and BH-FDR $q$-value across the 100 TV tests.}\label{tab:per-setup-tv-fdr}\\
\toprule
Dataset & Model & Regime & Stage & Break & $\overline{\Delta}$ & 95\% CI & $n$ & $p$ (two-sided) & $q$ (BH) \\
\midrule
\endfirsthead
\toprule
Dataset & Model & Regime & Stage & Break & $\overline{\Delta}$ & 95\% CI & $n$ & $p$ (two-sided) & $q$ (BH) \\
\midrule
\endhead
\midrule
\multicolumn{10}{r}{\emph{continued on next page}}\\
\midrule
\endfoot
\bottomrule
\endlastfoot
cifar100 & mobilenetv2 & negative & early & no & 0.000805 & [0.000750, 0.000858] & 672 & 0.00e+00 & 0.00e+00 \\
cifar100 & mobilenetv2 & negative & early & br & 0.000887 & [0.000799, 0.000974] & 352 & 0.00e+00 & 0.00e+00 \\
cifar100 & mobilenetv2 & orthogonal & early & no & 0.019348 & [0.017077, 0.021519] & 1280 & 0.00e+00 & 0.00e+00 \\
cifar100 & mobilenetv2 & orthogonal & early & br & -0.051217 & [-0.054100, -0.048418] & 896 & 0.00e+00 & 0.00e+00 \\
cifar100 & mobilenetv2 & resonant mid & early & no & 0.012552 & [0.010594, 0.014415] & 1280 & 0.00e+00 & 0.00e+00 \\
cifar100 & mobilenetv2 & resonant mid & early & br & -0.047301 & [-0.049171, -0.045360] & 1280 & 0.00e+00 & 0.00e+00 \\
cifar100 & mobilenetv2 & resonant strong & early & no & 0.041931 & [0.039873, 0.044059] & 1280 & 0.00e+00 & 0.00e+00 \\
cifar100 & mobilenetv2 & resonant strong & early & br & -0.027104 & [-0.029174, -0.025083] & 1280 & 0.00e+00 & 0.00e+00 \\
cifar100 & mobilenetv2 & standard & early & no & 0.015456 & [0.014300, 0.016628] & 1280 & 0.00e+00 & 0.00e+00 \\
cifar100 & mobilenetv2 & standard & early & br & -0.036990 & [-0.038282, -0.035757] & 1280 & 0.00e+00 & 0.00e+00 \\
cifar100 & resnet18 & negative & early & no & 0.000002 & [0.000002, 0.000003] & 320 & 0.00e+00 & 0.00e+00 \\
cifar100 & resnet18 & negative & early & br & 0.000002 & [0.000002, 0.000003] & 320 & 0.00e+00 & 0.00e+00 \\
cifar100 & resnet18 & orthogonal & early & no & -0.041033 & [-0.049857, -0.032092] & 640 & 0.00e+00 & 0.00e+00 \\
cifar100 & resnet18 & orthogonal & early & br & -0.025690 & [-0.034570, -0.016585] & 640 & 8.42e-08 & 9.91e-08 \\
cifar100 & resnet18 & resonant mid & early & no & -0.102151 & [-0.109609, -0.094925] & 640 & 0.00e+00 & 0.00e+00 \\
cifar100 & resnet18 & resonant mid & early & br & -0.109642 & [-0.117547, -0.102230] & 640 & 0.00e+00 & 0.00e+00 \\
cifar100 & resnet18 & resonant strong & early & no & -0.078765 & [-0.086192, -0.071730] & 640 & 0.00e+00 & 0.00e+00 \\
cifar100 & resnet18 & resonant strong & early & br & -0.090973 & [-0.098795, -0.083216] & 640 & 0.00e+00 & 0.00e+00 \\
cifar100 & resnet18 & standard & early & no & 0.015820 & [0.012206, 0.019341] & 640 & 0.00e+00 & 0.00e+00 \\
cifar100 & resnet18 & standard & early & br & 0.004149 & [-0.000556, 0.008643] & 640 & 8.12e-02 & 8.45e-02 \\
cifar100 & smallcnn & negative & early & no & 0.000000 & [0.000000, 0.000000] & 320 & 1.30e-07 & 1.52e-07 \\
cifar100 & smallcnn & negative & early & br & 0.000000 & [0.000000, 0.000000] & 320 & 5.35e-14 & 6.86e-14 \\
cifar100 & smallcnn & orthogonal & early & no & 0.032157 & [0.020628, 0.043640] & 640 & 4.04e-08 & 4.81e-08 \\
cifar100 & smallcnn & orthogonal & early & br & 0.011148 & [-0.000787, 0.022086] & 640 & 6.42e-02 & 6.76e-02 \\
cifar100 & smallcnn & resonant mid & early & no & -0.072155 & [-0.080590, -0.063648] & 640 & 0.00e+00 & 0.00e+00 \\
cifar100 & smallcnn & resonant mid & early & br & -0.055786 & [-0.065107, -0.046302] & 640 & 0.00e+00 & 0.00e+00 \\
cifar100 & smallcnn & resonant strong & early & no & -0.051914 & [-0.062084, -0.041965] & 640 & 0.00e+00 & 0.00e+00 \\
cifar100 & smallcnn & resonant strong & early & br & -0.039207 & [-0.048721, -0.029565] & 640 & 1.40e-14 & 1.82e-14 \\
cifar100 & smallcnn & standard & early & no & -0.008411 & [-0.011692, -0.004860] & 640 & 2.89e-06 & 3.25e-06 \\
cifar100 & smallcnn & standard & early & br & 0.009618 & [0.006112, 0.013103] & 640 & 1.98e-07 & 2.27e-07 \\
cifar100 & vgg11 & negative & early & no & 0.000000 & [0.000000, 0.000000] & 320 & 7.86e-13 & 9.71e-13 \\
cifar100 & vgg11 & negative & early & br & 0.000000 & [0.000000, 0.000000] & 320 & 9.08e-05 & 9.98e-05 \\
cifar100 & vgg11 & orthogonal & early & no & -0.021196 & [-0.023877, -0.018546] & 640 & 0.00e+00 & 0.00e+00 \\
cifar100 & vgg11 & orthogonal & early & br & -0.067943 & [-0.070836, -0.065066] & 640 & 0.00e+00 & 0.00e+00 \\
cifar100 & vgg11 & resonant mid & early & no & -0.001379 & [-0.003891, 0.000891] & 640 & 2.56e-01 & 2.64e-01 \\
cifar100 & vgg11 & resonant mid & early & br & -0.066817 & [-0.068861, -0.064791] & 640 & 0.00e+00 & 0.00e+00 \\
cifar100 & vgg11 & resonant strong & early & no & 0.033132 & [0.030105, 0.035919] & 640 & 0.00e+00 & 0.00e+00 \\
cifar100 & vgg11 & resonant strong & early & br & -0.049203 & [-0.051429, -0.046999] & 640 & 0.00e+00 & 0.00e+00 \\
cifar100 & vgg11 & standard & early & no & 0.011491 & [0.010437, 0.012573] & 640 & 0.00e+00 & 0.00e+00 \\
cifar100 & vgg11 & standard & early & br & -0.029600 & [-0.030528, -0.028720] & 640 & 0.00e+00 & 0.00e+00 \\
cifar100 & vit\_b16 & negative & early & no & 0.000000 & [0.000000, 0.000000] & 320 & 0.00e+00 & 0.00e+00 \\
cifar100 & vit\_b16 & negative & early & br & 0.000000 & [0.000000, 0.000000] & 320 & 0.00e+00 & 0.00e+00 \\
cifar100 & vit\_b16 & orthogonal & early & no & 0.016780 & [0.015028, 0.018559] & 640 & 0.00e+00 & 0.00e+00 \\
cifar100 & vit\_b16 & orthogonal & early & br & -0.062900 & [-0.064360, -0.061424] & 640 & 0.00e+00 & 0.00e+00 \\
cifar100 & vit\_b16 & resonant mid & early & no & 0.026613 & [0.025079, 0.028070] & 640 & 0.00e+00 & 0.00e+00 \\
cifar100 & vit\_b16 & resonant mid & early & br & -0.044202 & [-0.045560, -0.042761] & 640 & 0.00e+00 & 0.00e+00 \\
cifar100 & vit\_b16 & resonant strong & early & no & 0.055668 & [0.053859, 0.057515] & 640 & 0.00e+00 & 0.00e+00 \\
cifar100 & vit\_b16 & resonant strong & early & br & -0.029075 & [-0.030643, -0.027431] & 640 & 0.00e+00 & 0.00e+00 \\
cifar100 & vit\_b16 & standard & early & no & 0.012572 & [0.011751, 0.013416] & 640 & 0.00e+00 & 0.00e+00 \\
cifar100 & vit\_b16 & standard & early & br & -0.019568 & [-0.020261, -0.018862] & 640 & 0.00e+00 & 0.00e+00 \\
imagenette & mobilenetv2 & negative & early & no & 0.005157 & [0.004957, 0.005360] & 1280 & 0.00e+00 & 0.00e+00 \\
imagenette & mobilenetv2 & negative & early & br & 0.004266 & [0.004041, 0.004503] & 640 & 0.00e+00 & 0.00e+00 \\
imagenette & mobilenetv2 & orthogonal & early & no & 0.056586 & [0.055054, 0.058130] & 1280 & 0.00e+00 & 0.00e+00 \\
imagenette & mobilenetv2 & orthogonal & early & br & 0.007169 & [0.005253, 0.009030] & 640 & 5.86e-13 & 7.33e-13 \\
imagenette & mobilenetv2 & resonant mid & early & no & 0.060175 & [0.058912, 0.061466] & 1280 & 0.00e+00 & 0.00e+00 \\
imagenette & mobilenetv2 & resonant mid & early & br & 0.015760 & [0.014301, 0.017103] & 1024 & 0.00e+00 & 0.00e+00 \\
imagenette & mobilenetv2 & resonant strong & early & no & 0.072772 & [0.071373, 0.074186] & 1280 & 0.00e+00 & 0.00e+00 \\
imagenette & mobilenetv2 & resonant strong & early & br & 0.018428 & [0.017107, 0.019800] & 1280 & 0.00e+00 & 0.00e+00 \\
imagenette & mobilenetv2 & standard & early & no & 0.043493 & [0.042607, 0.044306] & 1280 & 0.00e+00 & 0.00e+00 \\
imagenette & mobilenetv2 & standard & early & br & 0.012223 & [0.011316, 0.013138] & 1280 & 0.00e+00 & 0.00e+00 \\
imagenette & resnet18 & negative & early & no & 0.000003 & [0.000003, 0.000004] & 320 & 0.00e+00 & 0.00e+00 \\
imagenette & resnet18 & negative & early & br & 0.000003 & [0.000003, 0.000004] & 320 & 0.00e+00 & 0.00e+00 \\
imagenette & resnet18 & orthogonal & early & no & -0.033066 & [-0.040169, -0.025908] & 640 & 0.00e+00 & 0.00e+00 \\
imagenette & resnet18 & orthogonal & early & br & -0.040914 & [-0.049604, -0.032405] & 640 & 0.00e+00 & 0.00e+00 \\
imagenette & resnet18 & resonant mid & early & no & -0.097785 & [-0.104683, -0.090893] & 640 & 0.00e+00 & 0.00e+00 \\
imagenette & resnet18 & resonant mid & early & br & -0.091842 & [-0.099027, -0.084716] & 640 & 0.00e+00 & 0.00e+00 \\
imagenette & resnet18 & resonant strong & early & no & -0.066178 & [-0.072768, -0.059103] & 640 & 0.00e+00 & 0.00e+00 \\
imagenette & resnet18 & resonant strong & early & br & -0.087421 & [-0.095146, -0.079758] & 640 & 0.00e+00 & 0.00e+00 \\
imagenette & resnet18 & standard & early & no & 0.016140 & [0.011878, 0.020230] & 640 & 3.02e-13 & 3.82e-13 \\
imagenette & resnet18 & standard & early & br & -0.022495 & [-0.027069, -0.018019] & 640 & 0.00e+00 & 0.00e+00 \\
imagenette & smallcnn & negative & early & no & 0.000000 & [0.000000, 0.000000] & 320 & 0.00e+00 & 0.00e+00 \\
imagenette & smallcnn & negative & early & br & 0.000000 & [0.000000, 0.000000] & 320 & 0.00e+00 & 0.00e+00 \\
imagenette & smallcnn & orthogonal & early & no & 0.002243 & [-0.006734, 0.011739] & 640 & 6.22e-01 & 6.22e-01 \\
imagenette & smallcnn & orthogonal & early & br & 0.011140 & [0.002560, 0.020306] & 640 & 1.42e-02 & 1.54e-02 \\
imagenette & smallcnn & resonant mid & early & no & -0.056493 & [-0.063667, -0.049287] & 640 & 0.00e+00 & 0.00e+00 \\
imagenette & smallcnn & resonant mid & early & br & -0.053091 & [-0.060871, -0.044799] & 640 & 0.00e+00 & 0.00e+00 \\
imagenette & smallcnn & resonant strong & early & no & -0.032882 & [-0.040282, -0.025792] & 640 & 0.00e+00 & 0.00e+00 \\
imagenette & smallcnn & resonant strong & early & br & -0.040571 & [-0.048600, -0.032944] & 640 & 0.00e+00 & 0.00e+00 \\
imagenette & smallcnn & standard & early & no & 0.001042 & [-0.001016, 0.003169] & 640 & 3.49e-01 & 3.55e-01 \\
imagenette & smallcnn & standard & early & br & 0.009399 & [0.006702, 0.012158] & 640 & 2.92e-11 & 3.57e-11 \\
imagenette & vgg11 & negative & early & no & 0.000000 & [0.000000, 0.000000] & 320 & 1.34e-06 & 1.52e-06 \\
imagenette & vgg11 & negative & early & br & 0.000000 & [0.000000, 0.000000] & 320 & 1.28e-08 & 1.54e-08 \\
imagenette & vgg11 & orthogonal & early & no & 0.032065 & [0.028080, 0.035953] & 640 & 0.00e+00 & 0.00e+00 \\
imagenette & vgg11 & orthogonal & early & br & 0.015976 & [0.012318, 0.019872] & 640 & 4.44e-16 & 5.84e-16 \\
imagenette & vgg11 & resonant mid & early & no & 0.018407 & [0.015339, 0.021351] & 640 & 0.00e+00 & 0.00e+00 \\
imagenette & vgg11 & resonant mid & early & br & 0.001541 & [-0.001617, 0.004896] & 640 & 3.51e-01 & 3.55e-01 \\
imagenette & vgg11 & resonant strong & early & no & 0.045932 & [0.042310, 0.049204] & 640 & 0.00e+00 & 0.00e+00 \\
imagenette & vgg11 & resonant strong & early & br & 0.020064 & [0.016778, 0.023461] & 640 & 0.00e+00 & 0.00e+00 \\
imagenette & vgg11 & standard & early & no & 0.002099 & [0.001116, 0.003087] & 640 & 4.08e-05 & 4.54e-05 \\
imagenette & vgg11 & standard & early & br & -0.005895 & [-0.006964, -0.004909] & 640 & 0.00e+00 & 0.00e+00 \\
imagenette & vit\_b16 & negative & early & no & 0.000000 & [0.000000, 0.000000] & 320 & 0.00e+00 & 0.00e+00 \\
imagenette & vit\_b16 & negative & early & br & 0.000000 & [0.000000, 0.000000] & 320 & 0.00e+00 & 0.00e+00 \\
imagenette & vit\_b16 & orthogonal & early & no & 0.039317 & [0.036467, 0.042267] & 640 & 0.00e+00 & 0.00e+00 \\
imagenette & vit\_b16 & orthogonal & early & br & -0.003246 & [-0.006069, -0.000244] & 640 & 2.97e-02 & 3.19e-02 \\
imagenette & vit\_b16 & resonant mid & early & no & 0.035099 & [0.032329, 0.037743] & 640 & 0.00e+00 & 0.00e+00 \\
imagenette & vit\_b16 & resonant mid & early & br & 0.002902 & [0.000042, 0.005710] & 640 & 4.13e-02 & 4.39e-02 \\
imagenette & vit\_b16 & resonant strong & early & no & 0.053741 & [0.050745, 0.056677] & 640 & 0.00e+00 & 0.00e+00 \\
imagenette & vit\_b16 & resonant strong & early & br & 0.015032 & [0.012409, 0.017768] & 640 & 0.00e+00 & 0.00e+00 \\
imagenette & vit\_b16 & standard & early & no & 0.017612 & [0.016491, 0.018653] & 640 & 0.00e+00 & 0.00e+00 \\
imagenette & vit\_b16 & standard & early & br & 0.006220 & [0.004754, 0.007658] & 640 & 2.22e-16 & 2.96e-16 \\
\end{longtable}
\end{landscape}

\section{QUANTITATIVE DOSE–RESPONSE (momentum, fixed aug$_B$).}
We model the predicted amplification due to momentum as
$A_\mu=\frac{1-\mu^k}{1-\mu}$ and regress the observed back–flow as
$\Delta = \alpha + \beta A_\mu + \gamma \rho + \varepsilon$,
where $\rho$ is the $A$–$B$ batch overlap.
Using TV under \emph{no break} and configurations with $\text{aug}_B=\texttt{weak}$
(\texttt{standard}: $k{=}3,\mu{=}0.90,\rho{=}0.5$;
\texttt{resonant\_mid}: $k{=}6,\mu{=}0.95,\rho{=}0.75$;
\texttt{resonant\_strong}: $k{=}6,\mu{=}0.99,\rho{=}1.0$),
we obtain $n=30$ dataset$\times$model points with $R^2=0.077$,
$\beta=-0.0273\,(0.0184)$, $p=1.38{\times}10^{-1}$ and
$\gamma=0.1607\,(0.1235)$, $p=1.93{\times}10^{-1}$.
Because $k$ and $\rho$ vary across regimes, we also perform a \emph{within-pair}
test at fixed $k{=}6$ (same dataset and model), comparing \texttt{resonant\_strong}
to \texttt{resonant\_mid}:
all $10/10$ pairs exhibit an increase (strong $>$ mid), with mean lift
$0.0251$ and 95\% CI $[0.0208, 0.0288]$ (paired $t$-test $p=8.14\times 10^{-7}$).
Numerically, for $k{=}6$ the amplification factor increases from
$A_\mu(\mu{=}0.95)\!\approx\!5.29$ to $A_\mu(\mu{=}0.99)\!\approx\!5.90$;
the joint increase in $A_\mu$ and $\rho$ yields a robust \emph{dose–response}
in $\Delta$.

\begin{figure}[h]
  \centering
  \includegraphics[width=.58\linewidth]{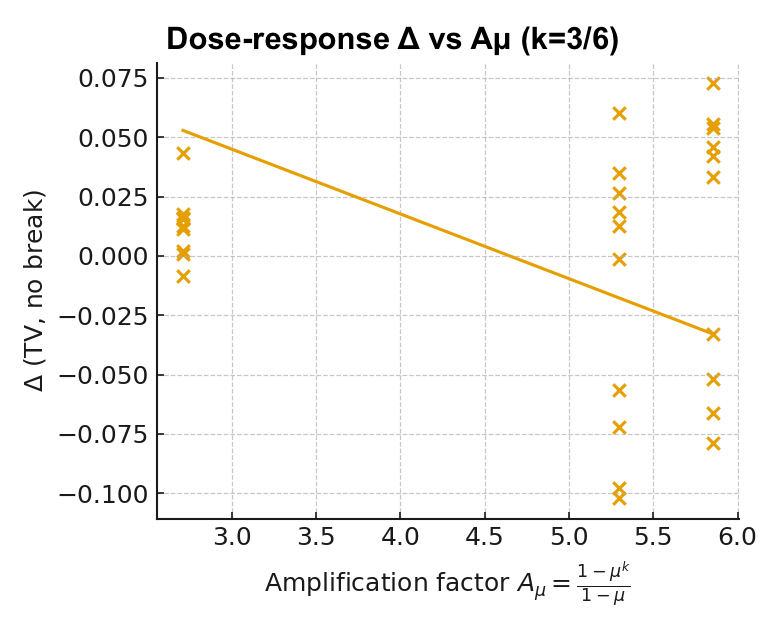}
  \caption{\textbf{Dose–response:} TV $\Delta$ vs $A_\mu=(1-\mu^k)/(1-\mu)$
  under \emph{no break} for regimes with $\text{aug}_B=\texttt{weak}$.
  Line shows OLS fit at mean overlap. A paired comparison at fixed $k{=}6$
  (not shown) finds strong$>$mid in $10/10$ dataset$\times$model pairs (mean lift $0.0251$, 95\% CI $[0.0208, 0.0288]$).}
  \label{fig:dose-response}
\end{figure}

\vfill

\end{document}